%% file: ms.tex
\documentclass[journal,onecolumn,11pt]{IEEEtran}
\usepackage{amsmath,amssymb,amsthm}
\usepackage{graphicx}
\usepackage{booktabs,longtable}
\usepackage{cite}
\usepackage[hidelinks]{hyperref}

\newtheorem{proposition}{Proposition}
\newtheorem{lemma}{Lemma}
\newtheorem{corollary}{Corollary}
\newtheorem{propS}[proposition]{Proposition}
\theoremstyle{remark}

\newcommand{\Pois}{\operatorname{Pois}}
\newcommand{\E}{\mathbb{E}}

\begin{document}

\title{Exponential Wake-Up Overshoot in Explicit\\ Gaussian Point-Process Filters}

\author{Hanwu~Qin and Dianhai~Wang%
\thanks{This work has been submitted to the IEEE for possible publication. Copyright may be transferred without notice, after which this version may no longer be accessible.}%
\thanks{Hanwu Qin and Dianhai Wang are with the Institute of Intelligent
Transportation Systems, Zhejiang University, Hangzhou 310058, China
(e-mails: hanwuqin@gmail.com; dianhaiwang@zju.edu.cn).}%
\thanks{Corresponding author: Hanwu Qin.}}

\markboth{Preprint}{Qin and Wang: Exponential Wake-Up Overshoot in Explicit Gaussian Point-Process Filters}

\maketitle

\begin{abstract}
When events resume after a long silence, explicit Gaussian point-process
filters amplify the first update exponentially in the silence duration. We
give the exact distribution of this update and a computable index that predicts
when it overshoots. For the explicit (single-scoring-step) filter
and its implicit posterior-mode counterpart, we derive second-order silence
expansions of the state variance with opposite-sign logarithmic corrections.
The crossing index determines when the post-wake prediction exceeds the resumed
rate, separating the bin width at which the amplification exponent reaches one
from the larger width at which the true rate is overshot. At fixed bin width,
the implicit filter's expected loss grows logarithmically with silence
duration, whereas a continuous-time counterexample rules out a bound uniform in
bin width and silence duration. Experimental evaluation confirms these
predictions in all 73 crossing cells and on exchange trade data, while variance
caps reduce explicit overshoot but mostly hurt the implicit filter.
\end{abstract}

\begin{IEEEkeywords}
Extended Kalman filter, implicit update, numerical stability, point processes,
state-space models.
\end{IEEEkeywords}

\IEEEpeerreviewmaketitle

\section{Introduction}

\IEEEPARstart{G}{aussian} point-process filters \cite{Snyder1972}, such as the
stochastic state point-process filter (SSPPF) \cite{Eden2004} and its
posterior-mode variant \cite{Smith2003}, accumulate state variance across
zero-count bins during a long silence, so the first resumed event is applied
with a large gain. Such filters track event rates in neural decoding
\cite{Brown1998,Truccolo2005}; this letter also analyzes them on exchange trade
streams, which arrive at irregular intervals \cite{Engle1998}. Because the
SSPPF linearizes the log-likelihood at the prior mean, it performs a single
Fisher-scoring step of the extended Kalman filter for dynamic GLMs
\cite{Fahrmeir1992}, whereas iterating to convergence yields the posterior-mode
(implicit) update \cite{Smith2003,Koyama2010,Lange2024}, as iterated Kalman
updates do for nonlinear measurements \cite{Bell1993,GarciaFernandez2015}.
Analyses of score-driven filters \cite{Lange2026,DonkerVanHeel2025}, a class
introduced in \cite{Creal2013}, and of implicit stochastic gradients
\cite{Toulis2014,Toulis2017} show that the explicit Poisson update can diverge
while the implicit update stays bounded by the likelihood peak. These works
treat the divergence as a property of the update rule; none quantifies the
silence-driven mechanism: how large the wake-up overshoot is, when it occurs, how it scales
with silence length and bin width, and what predictive risk the implicit update
retains. The effect is large: in our experiment, the median
excess loss in cells predicted to cross was $5.9\times10^{40}$ nats for the
explicit filter and $1.4\times10^{3}$ nats for the implicit one. In a
linear-measurement intensity observer, silence instead contracts the covariance
and is handled by dormant-state gating \cite{Qin2026}; with the log link
analyzed here, silence inflates the variance.

We quantify this silence-to-wake regime:
\begin{itemize}
\item We derive second-order silence variance expansions for both updates
  (Proposition~\ref{prop:silence}).
\item We obtain the exact wake-up distribution, a crossing index, and the
  post-peak loss (Proposition~\ref{prop:wake}).
\item We prove that the implicit update has logarithmic fixed-bin risk, whereas
  no risk bound holds uniformly in bin width and silence length
  (Propositions~\ref{prop:implicit} and~\ref{prop:ct}).
\item We confirm these predictions in experiments on synthetic
  streams and exchange trade data (Section~\ref{sec:exp}).
\end{itemize}
We build on the standard update recursions and the Lambert-$W$ form of the
implicit step \cite[eq.~(C.1)]{Lange2026}. Full proofs are in Appendices~\ref{sec:silence}--\ref{sec:ct}.

\section{Model, Filters, and Scoring}

We define the state-space model, explicit and implicit updates, and excess
log-loss metric. The log intensity follows a random walk observed through
binned counts,
\begin{equation}
\theta_{k+1}=\theta_k+w_k,\quad w_k\sim N(0,q),\quad
C_k\mid\theta_k\sim\Pois(e^{\theta_k}\Delta),
\end{equation}
with bin width $\Delta$ and per-bin variance $q=q_c\Delta$. Given the Gaussian
state $(\theta,V)$, the prediction is $(\theta^-,V^-)=(\theta,V+q)$. The two
updates are
\begin{align}
\text{E:}\;& V^+=\frac{V^-}{1+V^-e^{\theta^-}\Delta},\quad
\theta^+=\theta^-+V^+(C-e^{\theta^-}\Delta),\label{eq:E}\\
\text{I:}\;& \theta^+=\theta^-+V^-(C-e^{\theta^+}\Delta),\quad
V^+=\frac{V^-}{1+V^-e^{\theta^+}\Delta}.\label{eq:I}
\end{align}
Each bin is processed as predict, score, update. The median readout
$\hat\lambda=e^{\theta^-}$ is scored before the count is used. Against an
oracle with rate $\Lambda$ and $m=\Lambda\Delta$, the excess log loss of one
bin is $r=\hat\lambda\Delta-m-C\log(\hat\lambda/\Lambda)$, with nonnegative
conditional mean $\Delta[\hat\lambda-\Lambda+\Lambda\log(\Lambda/\hat\lambda)]$.
We write $s=\Delta\hat\lambda$ for the predicted count. We use the median
readout because the lognormal mean $e^{\theta+V/2}$ eventually grows during a
long silence.

\section{Silence}

In silence, both filter variances grow linearly at the same rate, and their
difference grows logarithmically.

\begin{proposition}[Second-order silence law]\label{prop:silence}
Fix $q,\Delta>0$ and an initial state $(\theta_0,V_0)$. After $n$ zero-count bins,
\begin{equation}
V_n^{j}=\frac{qn}{3}+a_j\log n+b_j+O\!\left(\frac{\log n}{n}\right),
\end{equation}
with $a_E=-2q/15$ and $a_I=+2q/15$, and $\Delta e^{\theta_n}\sim 6/(qn^2)$ for both $j\in\{E,I\}$. The constants obey
$b_j=q\,B_j(V_0/q,\,q\Delta e^{\theta_0})$ and are given by absolutely convergent series.
\end{proposition}

\begin{IEEEproof}[Sketch]
With $x=q/V$, $y=\Delta e^{\theta}V^2/q$ and $u=xy(1+x)$, both recursions give
$\log(y_{n+1}/y_n)=2x-3u+O(x^2+u^2)$, so $y\to2/3$ and $V_{n+1}-V_n\to q/3$.
The two updates first differ at order $u^2$, through the posterior variance.
Expanding to second order, $y_n=2/3+k_jx_n+O(x_n^2)$ with
$k_E=-38/45$ and $k_I=-22/45$. Then $V_{n+1}-V_n=q/3\mp(2q/45)x_n+O(qx_n^2)$, and
$x_n\sim3/n$ yields the $\mp(2q/15)\log n$ terms.
\end{IEEEproof}

\begin{figure*}[!t]
\centering
\includegraphics[width=0.86\textwidth]{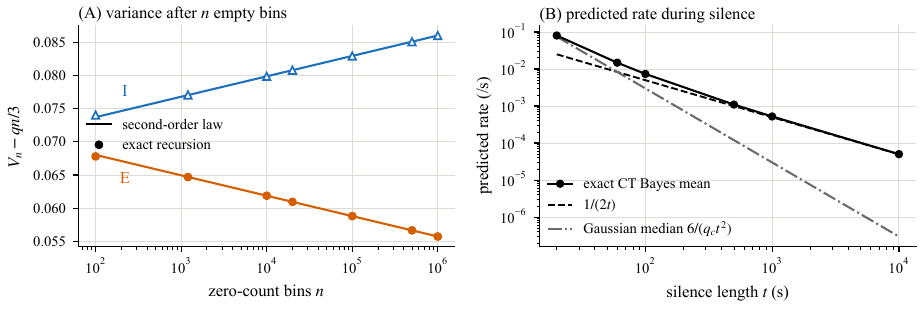}
\caption{Silence. (A) $V_n-qn/3$ for the exact recursions (dots) and the
second-order law of Proposition~\ref{prop:silence} (lines), with $b_j$ taken
from independent trajectories of length $10^6$; $q_c=0.2$/s, $\Delta=0.05$\,s,
initial rate $20$/s, $V_0=0.05$. (B) Exact CT Bayes predictive mean against
$1/(2t)$ and the Gaussian median readout $6/(q_ct^2)$, for the same $q_c$ and
initial state.}
\label{fig:silence}
\end{figure*}

The variances thus differ by $(4q/15)\log n$, whereas the median rates agree to
leading order (Fig.~\ref{fig:silence}A). In the continuous-time (CT) explicit
limit \cite{Eden2008}, $\dot\lambda=-W\lambda^2$ and $\dot W=q_c-W^2\lambda$
between events, with exact invariant $W^2=2q_c/(3\lambda)+K\lambda^2$ and
expansion $W(t)=q_ct/3+B+O(t^{-5})$. The Gaussian median decays like
$6/(q_ct^2)$, whereas the exact Bayes predictive mean decays like $1/(2t)$ by
the spectral representation of Brownian exponential functionals
\cite{Matsumoto2005} (Fig.~\ref{fig:silence}B), so after a long silence the
Gaussian median falls far below the exact predictive mean.

\section{Wake-Up}

Upon wake-up, the silence-grown variance acts as an exponential gain on the
count, governing overshoot and cooling.

\begin{proposition}[Exact wake-up update]\label{prop:wake}
Let the silent state be $(\theta_n,V_n)$, with $\lambda^-=e^{\theta_n}$,
$s=\Delta\lambda^-$, $v=V_n+q$, $g=v/(1+vs)$, $d=gs$, and $A=\lambda^-e^{-d}$. For
any count $C$, the explicit update gives
\begin{equation}
\lambda^+=A\,e^{gC}.\label{eq:wake}
\end{equation}
If $C\sim\Pois(m)$, the $p$-quantile of $\lambda^+$ is $Ae^{gQ_p(C)}$ and
$\E(\lambda^+)^r=A^r\exp\{m(e^{rg}-1)\}$ for $r>0$. The plug-in prediction
$\lambda^+$ for the next bin has expected excess loss, over the wake-up count
and the next count,
$se^{-d}\exp\{m(e^g-1)\}-m+m\{\log(m/s)+d-gm\}$.
\end{proposition}

By Proposition~\ref{prop:silence}, the gain $g\approx qn/3=q_cL/3$ in
\eqref{eq:wake} grows linearly with silence length $L=n\Delta$. For
$q_c=0.2$/s, $L=60$\,s, $\Delta=50$\,ms, initial rate 20/s and $V_0=0.05$, a
wake-up count $C\sim\Pois(50)$ gives a skewed post-wake rate with median
$10^{86.24}$/s and mean $10^{1245.2}$/s.

\begin{corollary}[Crossing index]\label{cor:cross}
At the mean count $C=m$, the explicit prediction exceeds $\Lambda$ if and only if
\begin{equation}
\kappa:=g\Lambda\Delta-\log(\Lambda/\lambda^-)-d>0.\label{eq:kappa}
\end{equation}
A risk level $\alpha$ is obtained by replacing $\Lambda\Delta$ with the
$(1-\alpha)$ Poisson quantile. When $m\ll1$, the first nonzero bin has $C\ge1$,
and the zero-truncated count must be used instead.
\end{corollary}

\begin{figure*}[!t]
\centering
\includegraphics[width=0.86\textwidth]{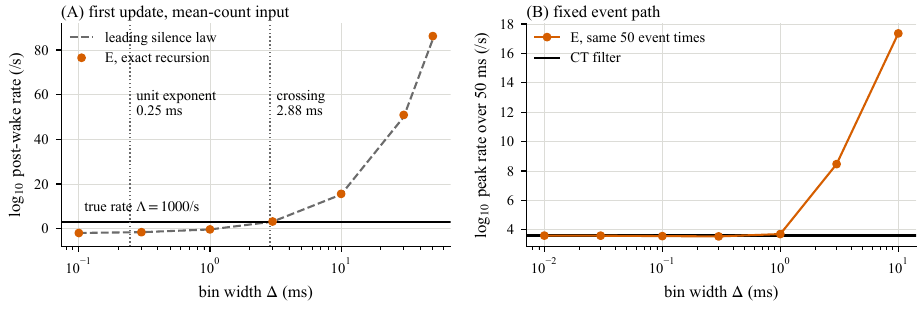}
\caption{Binning, from initial rate 20/s and $V_0=0.05$ with $q_c=0.2$/s and
$L=60$\,s. (A) Post-wake rate under a mean-count input for $\Lambda=1000$/s:
the unit-exponent width ($0.25$\,ms) and crossing width ($\approx2.88$\,ms)
differ by an order of magnitude. (B) For 50 fixed event times in the 50\,ms
after the silence, the discrete explicit filter converges to the event-aligned
CT filter as $\Delta\downarrow0$ (CT peak $4049$/s).}
\label{fig:binning}
\end{figure*}

With the leading silence law, the mean-count exponent
$g\Lambda\Delta\approx q_cL\Lambda\Delta/3$ reaches one at width
$3/(q_cL\Lambda)$, whereas crossing the true rate also requires closing the
silence gap $\log(\Lambda/\lambda^-)$. Under a mean-count input for
$q_c=0.2$/s, $L=60$\,s, and $\Lambda=1000$/s, the unit-exponent width is
$0.25$\,ms and the crossing width is about $2.88$\,ms (Fig.~\ref{fig:binning}A).
After a peak, \eqref{eq:E} gives, for the posterior $(\theta_j,V_j)$ and
$s_j=\Delta e^{\theta_j}$, $\theta_{j+1}-\theta_j=-1+\varepsilon_j$ with
$\varepsilon_j=(1+(V_j+q)C_j)/(1+(V_j+q)s_j)$, cooling by about one nat per
bin while $s_j\gg C_j+1/q$. Under the mean count input and
$\lambda_{\rm peak}\to\infty$ at fixed $q$ and $\Delta$, cooling takes
$\Delta\log(\lambda_{\rm peak}/\Lambda)+O(\Delta)$, with a constant that is not
uniform as $q\downarrow0$, and the excess loss is asymptotically
$\lambda_{\rm peak}\Delta\,e/(e-1)$.

\section{The Implicit Update}

At fixed bin width, the implicit update incurs logarithmic wake-up loss, while
in continuous time it shares the explicit filter's exponential silence
sensitivity. Lange et al.\ proved that the implicit posterior rate lies between
the prior and empirical rates; we build on this result.

\begin{lemma}[{[\citen{Lange2026}, Prop.~1]}]\label{lem:peak}
Equation~\eqref{eq:I} has a unique root. For $C>0$, $\lambda^+$ lies between
$\lambda^-$ and $C/\Delta$.
\end{lemma}

Because scoring precedes updating, the initial wake-up bin is predicted at the
silent rate of order $n^{-2}$, generating the implicit loss.

\begin{proposition}[Fixed-bin implicit risk]\label{prop:implicit}
Fix $\Delta,q_c,\Lambda>0$ and a window of $H$ bins after the wake-up, with
$C_j\sim\Pois(m)$ i.i.d. As $n\to\infty$,
\begin{equation}
\E\mathcal{R}^{I}_H=\frac{2m(1-e^{-Hm})}{1-e^{-m}}\log n+O(1).\label{eq:implog}
\end{equation}
\end{proposition}

\begin{IEEEproof}[Sketch]
Before the first positive count, the predicted log rate is
$-2\log n+O(1)$ (Proposition~\ref{prop:silence}). After it, the implicit rate
tends to $C/\Delta$ and the remaining bins contribute $O(1)$. The only divergent
term is $2C_J\log n$, where $J$ is the first bin with a positive count; its
factor 2 is the decay exponent of the median readout. Its expectation is $\sum_{j\le H}e^{-(j-1)m}m$. Dominated convergence follows from
a bound $K(1+N^2)$ on the window loss, where $N$ is the window count.
\end{IEEEproof}

\begin{proposition}[No joint bound]\label{prop:ct}
In the CT filter, the first event after a silence of length $L$ multiplies the
rate by $e^{W}$, where $W=q_cL/3+O(1)$. Under a resumed Poisson($\Lambda$)
stream, the expected excess loss over any fixed window is
$\Omega(e^{q_cL/9}/(q_cL))$. Both \eqref{eq:E} and \eqref{eq:I}
converge to this CT filter along fixed event paths as $\Delta\downarrow0$.
Hence no bound on the implicit risk holds uniformly in both $\Delta$ and $L$.
\end{proposition}

The implicit update is thus a fixed-bin remedy. In the CT limit, for
$q_c=0.2$/s and $L=600$\,s, a single event lifts the rate to $2\times10^{13}$/s.

\section{Experimental Evaluation}\label{sec:exp}

Experiments on synthetic and exchange streams evaluate the crossing index,
fixed-bin implicit risk, and variance capping under a prespecified protocol.

\emph{Protocol.} Methods, cells, tests, and thresholds were fixed before
formal synthetic-data generation and real-data test evaluation, and bound to a
protocol hash.
Descriptive summaries of the real test days had been inspected before the
experimental design was finalized, as disclosed in the protocol and Appendix~\ref{sec:protocol}. The design was amended once before
formal test evaluation: the real-data tuning grid was widened after the explicit
filter selected its lower edge on the training days. The \emph{silence--wake} (SW) streams have
300\,s at rate $\Lambda$, a silence of length $L$, and 120\,s at $\Lambda$, with
$\Lambda\in\{1,10,100,1000\}$/s, $L\in\{5,20,60,300,1200\}$\,s, and
$q_c\in\{0.02,0.2\}$/s. Continuous-time events were binned at
$\Delta\in\{1,10,50,250,1000\}$\,ms across 200 cells of 64 streams each. Excess
loss is summed after a 60\,s burn-in, including the silence. Before data
generation, a deterministic recursion (zero counts over the silence, then the
mean count), started from $\theta=\log\Lambda$ and the stationary E variance
(the filters' initial state), assigned each cell a frozen $\kappa$:
crossing if $\kappa\ge\log10$ (73 cells), no crossing if $\kappa\le-\log10$ (92
cells), transition otherwise (35 cells). The capped filters use
$V^-=\min(V+q,V_{\max})$ with $V_{\max}=1$. The control (CTRL) streams last
1800\,s without silence: the base rate oscillates between 1 and $1000$/s with
period 300\,s plus a three-state jump of $\pm0.5$ in log rate, with 32 streams
in each of the 10 cells ($\Delta\times q_c$; waveform in Appendix~\ref{sec:protocol}). Exact grid Bayes (X) was run at $\Delta=1$\,s.

The real data are Binance USD-M perpetual \texttt{aggTrades} for ten
stock- and index-linked contracts \cite{BinanceVision} (tickers and dates in
Appendix~\ref{sec:protocol}), with each trade treated as an
event and $\Delta\in\{10,50,250,1000\}$\,ms. The parameters $q_c$ and, for the
capped filters, $V_{\max}$ were tuned on a grid (Appendix~\ref{sec:protocol}) per method and bin
width on two training days and frozen; scores were computed on two
test days (20 contract-days), with bin score
$C\log(\hat\lambda\Delta)-\hat\lambda\Delta-\log C!$. The protocol prespecified
1\,s and 250\,ms for R1 and R2; 10 and 50\,ms are descriptive. Per-cell tests are exact one-sided sign tests on
paired streams, and the non-inferiority tests (H3, R3) are paired $t$ tests.
Holm correction is applied within each synthetic family at level $0.05/3$ and
across R1--R3 at level $0.05$.

\begin{table}[!t]
\caption{Confirmatory Results (Holm-Adjusted)}
\label{tab:res}
\centering
\begin{tabular}{@{}llcl@{}}
\toprule
& Hypothesis & Confirmed & Evidence\\
\midrule
H1 & I $<$ E, crossing cells & 73/73 & $p\le2.0\times10^{-8}$\\
H2 & both caps help$^\ast$ & 2/24 & I$_{\rm cap}$ fails\\
H3 & CTRL, non-inferior$^\dagger$ (2\%) & 10/10 & $p\le9.2\times10^{-12}$\\
R1 & real, 1\,s: I $>$ E & yes & 18:2, $p=6.0\times10^{-4}$\\
R2 & real, 250\,ms: I $>$ E & yes & 18:2, $p=6.0\times10^{-4}$\\
R3 & real: I$_{\rm cap}$ non-inferior (0.5\%) & no & $p=0.91$\\
\bottomrule
\end{tabular}\\[2pt]
\parbox{0.6\textwidth}{\footnotesize H rows compare excess loss (lower is
better); R rows compare log score (higher is better).
$^\ast$Cells with $\Delta\le10$\,ms and $q_cL/3\ge8$; both
$\mathrm{E_{cap}}<\mathrm{E}$ and $\mathrm{I_{cap}}<\mathrm{I}$ required.
$^\dagger$I against E, and I$_{\rm cap}$ against I; margins relative to the
comparator's loss (H3) or to $|\mathrm{score(I)}|$ (R3). Holm runs over 73, 24,
and 10 cells in H1--H3 (level $0.05/3$); H2 and H3 are intersection--union
tests; H1 and H3 show the largest adjusted $p$.}
\end{table}

\begin{figure}[!t]
\centering
\includegraphics[width=0.6\textwidth]{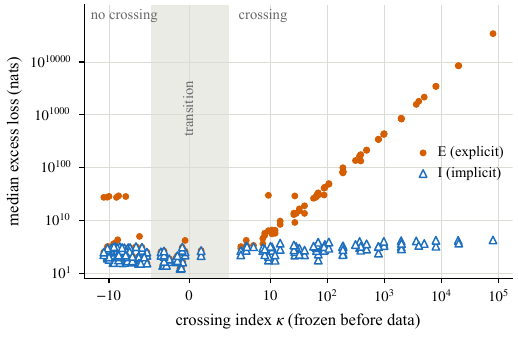}
\caption{Median excess loss per cell (64 streams) against the crossing index
$\kappa$ frozen before the data; the loss axis is doubly logarithmic. The
shaded band is the transition region. The
cluster of explicit losses near $10^{27}$ at $\kappa<0$ comes from $L=1200$\,s,
$q_c=0.2$/s, $\Delta\le50$\,ms, where $m\ll1$ and the mean-count index
under-predicts the crossing (Corollary~\ref{cor:cross}).}
\label{fig:exp}
\end{figure}

\emph{Results.} Table~\ref{tab:res} reports the confirmatory tests; the
remaining summary statistics are descriptive.

\emph{Explicit versus implicit.} In the crossing region, the initial post-wake
prediction of E exceeded $\Lambda$ in 92.96\% of the streams, against 1.89\% in
the no-crossing region. The median excess loss was $5.9\times10^{40}$\,nats
for E and $1.4\times10^{3}$\,nats for I. In 29.8\% of the crossing streams, E
failed to re-enter $[\Lambda/1.5,1.5\Lambda]$ for a full second within 120\,s,
whereas I re-entered in every stream with a median of 0.25\,s. In
Fig.~\ref{fig:exp}, the explicit median loss rises steeply across the
transition band, while every implicit cell median stays below
$2\times10^4$\,nats.

\emph{Exceptions to the frozen regions.} The exceptions are cells with $m\ll1$
and $q_cL/3=80$, where the mean-count proxy predicts no crossing, whereas a
single event gives $g-\log(\Lambda/\lambda^-)-d\approx70$ under the
zero-truncated count (Corollary~\ref{cor:cross}).

\emph{Logarithmic implicit risk.} Consistent with the fixed-bin law
\eqref{eq:implog}, the median implicit loss at $\Lambda=1000$/s and
$\Delta=1$\,s grew by 5870 and 5864\,nats from $L=60$ to $1200$\,s ($q_c=0.02$
and $0.2$/s), near the predicted increase $2m\log(1200/60)\approx5991$\,nats;
the added silent bins contribute under 5\,nats. At the same width
and rate, the median paired advantage of exact Bayes over I was
$8.7\times10^3$ to $1.7\times10^4$\,nats for $L\ge60$\,s, consistent with the
gap between the Gaussian median and the exact mean in Fig.~\ref{fig:silence}B.

\emph{Variance caps.} Capping reduced E's loss in every H2 cell with
$q_c=0.2$/s, where E$_{\rm cap}$ beat E in 64 of 64 streams; in the crossing
region its median loss was still $4.5\times10^7$\,nats. Across the H2 cells,
I$_{\rm cap}$ lost 1336 of the 1536 paired comparisons against I, and H2 held
only at $\Lambda=1$/s and $\Delta=1$\,ms, near the CT regime of
Proposition~\ref{prop:ct}.

\emph{Real data.} At 1\,s, I beat E by $6.4\times10^4$\,nats over the 20 test
contract-days (2.5\% of the total log loss of E); I also beat E on 20 of 20
contract-days at 10\,ms and 18 of 20 at 50\,ms. Because contract-days share
market conditions, we also summed each contract over its two test days
(exploratory): I beat E for 10 of 10 contracts at 1\,s and 9 of 10 at 250\,ms
(sign tests, $p=9.8\times10^{-4}$ and $0.011$). Non-inferiority of
I$_{\rm cap}$ to I was not confirmed (R3), and with a tuned $V_{\max}$,
I$_{\rm cap}$ again had a lower total score than I at 50 to 1000\,ms. Across all bin widths, E produced 13
bins (at 10 and 50\,ms) whose predicted rate exceeded 100 times the
contract-day's maximum one-second count; I produced none.

\section{Conclusion}

Long silences turn the explicit Gaussian point-process filter into an
exponential amplifier with gain set by the silent state. In practice, the
crossing index~\eqref{eq:kappa}, evaluated at the mean count or the
zero-truncated count when $m\ll1$, identifies when the explicit update
overshoots. At any fixed bin width, the implicit update confines post-wake loss
to logarithmic growth, whereas shrinking the bin width toward continuous time
restores exponential growth in the silence length. Variance capping reduces
the overshoot of the explicit update, whereas capping the implicit update
mostly increased its loss in our experiments.

\clearpage
\appendices
\noindent Appendix~\ref{sec:conv} fixes the conventions, Appendices~\ref{sec:silence}--\ref{sec:ct}
give the proofs, and Appendices~\ref{sec:protocol} and~\ref{sec:results} document the
experimental protocol and additional results. Equations, results and tables continue the
numbering of the main text.

\section{Conventions}\label{sec:conv}

This section fixes the filter notation, loss definitions, and asymptotic regimes
used throughout the proofs. The state model, the explicit update E, and the
implicit update I are those of (1)--(3) in the main text. Throughout,
$q=q_c\Delta$, $n$ is the number of consecutive zero-count bins, $L=n\Delta$,
and $(\theta_n,V_n)$ is the posterior after them. The predicted count is
$s=\Delta e^{\theta^-}$ and the predicted variance is $v=V+q$; E and I apply no
variance cap, exponent guard, or readout clipping unless a capped filter is
named.

Each bin is processed as predict, score, update. Against an oracle of rate
$\Lambda$ and $m=\Lambda\Delta$, the realized excess loss of one bin is
\begin{equation}
r=s-m-C\log(s/m),\label{eq:loss}
\end{equation}
which can be negative, whereas its conditional mean given the past is the
Kullback--Leibler divergence
\begin{equation}
\E[r\mid\mathcal F]=\Delta\{\hat\lambda-\Lambda+\Lambda\log(\Lambda/\hat\lambda)\}
=:d_\Delta(\Lambda,\hat\lambda)\ge0 .\label{eq:kl}
\end{equation}
Because the prediction is predictable, the expected sum of realized losses
equals the expected sum of conditional divergences whenever both are
integrable.

The proofs keep two limits apart. Proposition~1 and
Proposition~3 fix $q$ (or $\Delta$) and let $n\to\infty$, whereas the binning
results fix $q_c$ and $L$ and take the joint limit $\Delta\downarrow0$
($q\downarrow0$, $n\uparrow\infty$). No result below exchanges these limits.

\section{Silence}\label{sec:silence}

This section proves Proposition~1 in the refined form of
Proposition~\ref{prop:S-silence}. It then derives the wake gain
(Corollary~\ref{cor:S-gain}) and the continuous-time silence limit with its exact
invariant (Proposition~\ref{prop:S-ct}), and shows that under exact Bayes in
silence the discrete predictive means are not summable
(Proposition~\ref{prop:S-nonsum}) while the continuous-time predictive mean
decays algebraically (Proposition~\ref{prop:S-bayes}).

\subsection{Proof of Proposition 1}

We prove the following, slightly fuller, statement.

\begin{propS}\label{prop:S-silence}
Fix $q,\Delta>0$, $V_0\ge0$ and $\theta_0\in\mathbb R$, and write
$s_n=\Delta e^{\theta_n}$. For $j\in\{E,I\}$ there are finite constants $b_j$ with
\begin{equation}
V_n^j=\frac{qn}{3}+a_j\log n+b_j+O\!\left(\frac{\log n}{n}\right),\qquad
a_E=-\frac{2q}{15},\quad a_I=\frac{2q}{15},\label{eq:S-V}
\end{equation}
\begin{equation}
\log s_n^j=\log\frac{6}{qn^2}
+\frac{-(6a_j/q)\log n-6b_j/q+\gamma_j}{n}
+O\!\left(\frac{\log^2n}{n^2}\right),\qquad
\gamma_E=-\frac{19}{5},\quad\gamma_I=-\frac{11}{5}.\label{eq:S-s}
\end{equation}
The constants are given by the absolutely convergent series
\begin{equation}
b_j=V_0+\sum_{k=0}^{\infty}\Bigl[V^j_{k+1}-V^j_k-\frac q3-a_j\log\frac{k+2}{k+1}\Bigr],
\label{eq:S-b}
\end{equation}
and satisfy the exact scaling $b_j=q\,B_j(V_0/q,\,q\Delta e^{\theta_0})$.
\end{propS}

The scaling holds because, in the variables $V_n/q$ and $qs_n$, neither
recursion contains $q$. Truncating~\eqref{eq:S-b} at $N$ is equivalent to
$b_{j,N}=V_N^j-qN/3-a_j\log(N+1)$, with error $O(\log N/N)$. No closed form of
$b_j$ for general initial values is claimed.

\begin{proof}
\emph{Step 1: entry into the asymptotic regime.} Put $A_n=V_n+q$ and
$\rho_n=A_ns_n$. A zero-count E step is
\[
d_n=\frac{\rho_n}{1+\rho_n},\qquad V_{n+1}=\frac{A_n}{1+\rho_n},\qquad s_{n+1}=s_ne^{-d_n},
\]
and a zero-count I step is the same with $d_n=W_0(\rho_n)$ and
$V_{n+1}=A_n/(1+d_n)$, where $W_0$ is the principal Lambert function. In both
cases $s_n$ decreases strictly. If its limit were positive, $A_n\ge q$ would
bound $d_n$ away from zero, a contradiction; hence $s_n\to0$.

For any $M>0$, eventually $V_n\le M$ implies $V_{n+1}\ge V_n+q/2$, while
$V_n\ge M$ implies $V_{n+1}>M$, because $A/(1+As)$ and $A/(1+W_0(As))$ increase
in $A$ and tend to $A$ as $s\to0$. Hence $V_n\to\infty$. Moreover
\[
\frac{\rho_{n+1}}{\rho_n}=\Bigl(f_j(\rho_n)+\frac q{A_n}\Bigr)e^{-d_n},\qquad
f_E(\rho)=\frac1{1+\rho},\quad f_I(\rho)=\frac1{1+W_0(\rho)} .
\]
For $\rho_n\ge\varepsilon$ the right side is eventually below a constant smaller
than one, and for $\rho_n<\varepsilon$ it is at most $1+q/A_n\to1$. Hence
$\rho_n\to0$.

Define
\[
x_n=\frac q{V_n},\qquad y_n=\frac{s_nV_n^2}{q},\qquad \rho_n=x_ny_n(1+x_n).
\]
Then $x_n\to0$, $\rho_n\to0$, and $\sum_nx_n=\infty$ because $V_n\le V_0+nq$.
Both recursions give
\begin{equation}
\log\frac{y_{n+1}}{y_n}=2x_n-3\rho_n+R_n,\qquad |R_n|\le K(x_n^2+\rho_n^2).
\label{eq:S-logy}
\end{equation}
Since $y_n$ is not yet known to be bounded, we compare against outer bands.
For any small $\delta>0$, eventually $|R_n|\le\delta(x_n+\rho_n)$, so
\[
(2-\delta)x_n-(3+\delta)\rho_n\le\log\frac{y_{n+1}}{y_n}\le(2+\delta)x_n-(3-\delta)\rho_n .
\]
Fix $0<\varepsilon<2/3$. Choosing $\delta$ small and then $n$ large, the right
side is at most $-c_\varepsilon x_n$ on the whole unbounded region
$y_n\ge2/3+\varepsilon$, and the left side is at least $c_\varepsilon x_n$ on
$0<y_n\le2/3-\varepsilon$, for some $c_\varepsilon>0$. The first comparison
holds for arbitrarily large $y_n$ because the remainder is controlled relative
to $x_n+\rho_n$ and not only to $x_n$. Since $\sum x_n=\infty$, the orbit cannot
stay in either outer band, so it enters $[2/3-\varepsilon,2/3+\varepsilon]$.
Because $\log(y_{n+1}/y_n)\to0$, the drift points inward next to that band, and
one step cannot jump across the enlarged band
$[2/3-2\varepsilon,2/3+2\varepsilon]$, the orbit eventually stays in the enlarged
band. Letting $\varepsilon\downarrow0$ gives $y_n\to2/3$, and therefore
\[
V_{n+1}-V_n=q(1-y_n+o(1))\to\frac q3,\qquad \frac{V_n}{n}\to\frac q3,\qquad s_n\sim\frac6{qn^2}.
\]

\emph{Step 2: second-order coefficients.} Near $y=2/3$, a Taylor expansion with
uniform remainder gives, with $\alpha_E=1$ and $\alpha_I=2$,
\begin{align}
\frac{V_{n+1}}{V_n}&=1+(1-y)x+(-2y+\alpha_jy^2)x^2+O(x^3),\label{eq:S-Vratio}\\
\log\frac{y_{n+1}}{y_n}&=(2-3y)x+(-1-3y+2\alpha_jy^2)x^2+O(x^3).\label{eq:S-yratio}
\end{align}
For E, \eqref{eq:S-Vratio} follows from $V_{n+1}/V_n=(1+x)/(1+\rho)$ with
$\rho=xy+x^2y$. For I it follows from $1/(1+W_0(\rho))=1-\rho+2\rho^2+O(\rho^3)$.
Writing $u_n=y_n-2/3$,
\begin{align*}
u_{n+1}&=u_n-2x_nu_n-3x_nu_n^2+D_jx_n^2+O(x_n^2|u_n|+x_n^3),\qquad
D_E=-\tfrac{38}{27},\quad D_I=-\tfrac{22}{27},\\
x_{n+1}&=x_n-\tfrac13x_n^2+u_nx_n^2+O(x_n^3).
\end{align*}
Here $D_j=\tfrac23(-3+\tfrac89\alpha_j)$ is the $x^2$ coefficient of
$y\exp\{(2-3y)x+\cdots\}$ at $y=2/3$. Since $x_n\sim3/n$ and $u_n\to0$, eventually
$|u_{n+1}|\le(1-5/n)|u_n|+K/n^2$. Comparison with $M/n$ for large $M$ gives
$u_n=O(x_n)$. Put $k_j=3D_j/5$, so that
\[
k_E=-\frac{38}{45},\qquad k_I=-\frac{22}{45},
\]
and $w_n=u_n-k_jx_n$. The $x_n^2$ terms cancel exactly, because
$D_j+k_j/3-2k_j=D_j-5k_j/3=0$, and
\[
w_{n+1}=(1-2x_n+O(x_n^2))w_n+O(x_n^3).
\]
Comparison with $M/n^2$ gives $w_n=O(n^{-2})$, hence
\begin{equation}
y_n=\frac23+k_jx_n+O(x_n^2).\label{eq:S-y}
\end{equation}

\emph{Step 3: summation.} Substituting~\eqref{eq:S-y} into~\eqref{eq:S-Vratio},
\[
V_{n+1}-V_n=\frac q3+A_jqx_n+O(qx_n^2),\qquad
A_j=-k_j-\frac43+\frac{4\alpha_j}9,\quad A_E=-\frac2{45},\quad A_I=\frac2{45}.
\]
A first summation gives $V_n=qn/3+O(\log n)$ and hence $x_n=3/n+O(\log n/n^2)$.
Then $R'_n=V_n-qn/3-3qA_j\log n$ satisfies $R'_{n+1}-R'_n=O(\log n/n^2)$,
which is absolutely summable with tail $O(\log n/n)$. This proves~\eqref{eq:S-V}
with $a_j=3qA_j$ and~\eqref{eq:S-b}. Finally, $s_n=qy_n/V_n^2$ together
with~\eqref{eq:S-y} and~\eqref{eq:S-V} gives~\eqref{eq:S-s}, with
$\gamma_j=9k_j/2$.
\end{proof}

\begin{corollary}[Wake gain]\label{cor:S-gain}
Under the assumptions of Proposition~\ref{prop:S-silence}, the gain of an explicit
wake-up update is
\[
g_n=\frac{V_n+q}{1+(V_n+q)s_n}=\frac{qn}3+a_E\log n+b_E+\frac q3+O\!\left(\frac{\log n}n\right),
\]
and for any count $C\ge0$,
\begin{equation}
\log(\Delta\lambda^+)=\log\frac6{qn^2}+\frac{qnC}3+C\Bigl[a_E\log n+b_E+\frac q3\Bigr]
+O\!\left((1+C)\frac{\log n}n\right).\label{eq:S-wake2}
\end{equation}
If $C=C_n$ grows with $n$, the remainder grows with it.
\end{corollary}

\begin{proof}
$g_n$ is the variance $V_{n+1}$ that one more zero-count step would produce, and
$\log(\Delta\lambda^+)=\log s_n+g_n(C-s_n)$ exactly
(Proposition~\ref{prop:S-wake}). Substitute~\eqref{eq:S-V} and~\eqref{eq:S-s}.
\end{proof}

\paragraph{Numerical check.} For $q=0.01$, $\Delta=0.05$\,s, $\lambda_0=20$/s
and $V_0=0.05$, the exact recursions give $V_{1200}^E=4.064702$ and
$V_{1200}^I=4.077060$. With $b_j$ estimated from independent trajectories of
length $10^6$ ($b_E\approx0.074174$, $b_I\approx0.067576$), the second-order law
has absolute error $1.9\times10^{-5}$ (E) and $3.1\times10^{-5}$ (I) at
$n=1200$. The expansion is not uniform in $q$: for $q=10^{-8}$ and $n=10^6$,
$V_n/(qn/3)$ is still $1.18$ for both updates.

\subsection{Continuous-time limit}

\begin{propS}[CT invariant]\label{prop:S-ct}
Let $q_c>0$, $\lambda(0)=\lambda_0>0$, $W(0)=W_0\ge0$, and no events. The CT
filter $\dot\lambda=-W\lambda^2$, $\dot W=q_c-W^2\lambda$ satisfies
\begin{equation}
W(t)^2=\frac{2q_c}{3\lambda(t)}+K\lambda(t)^2,\qquad
K=\frac{W_0^2-2q_c/(3\lambda_0)}{\lambda_0^2}.\label{eq:S-inv}
\end{equation}
With
\[
B=W_0-K\int_0^{\lambda_0}\frac{u^{3/2}}{\sqrt{2q_c/3+Ku^3}}\,du,
\]
we have $W(t)=q_ct/3+B+O(t^{-5})$ and $\lambda(t)=6/\{q_c(t+3B/q_c)^2\}+O(t^{-8})$.
If $\eta_t=|3K\lambda(t)^3/(2q_c)|<1$ and $A_t=\sqrt{2q_c(1-\eta_t)/3}$, then
\begin{equation}
\Bigl|W(t)-\frac{q_ct}3-B\Bigr|\le\frac{2|K|}{5A_t}\lambda(t)^{5/2}.\label{eq:S-ctbound}
\end{equation}
\end{propS}

\begin{proof}
Where $W>0$, eliminating $t$ gives $dW^2/d\lambda-2W^2/\lambda=-2q_c/\lambda^2$,
that is $d(W^2/\lambda^2)/d\lambda=-2q_c/\lambda^4$, whose integral
is~\eqref{eq:S-inv}; $W_0=0$ follows by continuity since $\dot W=q_c>0$ at
$W=0$. Thus $\lambda$ decreases strictly. A positive limit of $\lambda$ would
force $\dot\lambda\to0$ with finite $W$, which is incompatible with
$\dot W=q_c-W^2\lambda$; hence $\lambda\to0$ and $W\to\infty$.
By~\eqref{eq:S-inv}, $\dot W=q_c/3-K\lambda^3$, so
$W(t)-q_ct/3-B=K\int_t^\infty\lambda(u)^3du$, where the definition of $B$ uses
$dt=-d\lambda/(W\lambda^2)$. For $u\ge t$, $W(u)\ge A_t\lambda(u)^{-1/2}$, and
$\int_t^\infty\lambda^3du=\int_0^{\lambda(t)}z\,dz/W(z)\le2\lambda(t)^{5/2}/(5A_t)$,
which gives~\eqref{eq:S-ctbound}. Since $\lambda\sim6/(q_ct^2)$, the remainder is
$O(t^{-5})$, and solving~\eqref{eq:S-inv} for $\lambda$ gives the $O(t^{-8})$ term.
\end{proof}

For $q_c=0.2$/s, $\lambda_0=20$/s and $W_0=0.05$, $B=0.0748147$ and the time shift
is $3B/q_c=1.12222$\,s. At $t=60$\,s the shifted law has error
$6.3\times10^{-11}$ against a bound~\eqref{eq:S-ctbound} of $6.6\times10^{-11}$.

\subsection{Exact Bayes in silence}

The main text compares the Gaussian median with the exact predictive mean. The
following two statements show that the exact mean is not summable in discrete
time and decays like $1/(2t)$ in continuous time. The spectral representation
and the first-order survival constant are known results on exponential
functionals of Brownian motion~\cite{Matsumoto2005}; only their use for the
filtering readout is ours.

\begin{propS}[Discrete non-summability]\label{prop:S-nonsum}
Fix $\Delta,q>0$ and a Gaussian initial $\theta_0$. Let
$m_k=\E[e^{\theta_k}\mid C_1=\cdots=C_{k-1}=0]$. Then $\sum_km_k=\infty$. In
particular, no bound $m_k=O(k^{-1-\epsilon})$ holds, and the exact predictive
mean cannot be of the order $6/(q\Delta k^2)$ of the Gaussian median.
\end{propS}

\begin{proof}
Let $S_n=\Pr(C_1=\cdots=C_n=0)$. Decomposing by predictive distributions and
applying Jensen's inequality,
$S_n=\prod_{k\le n}\E[e^{-\Delta e^{\theta_k}}\mid C_{1:k-1}=0]\ge\exp(-\Delta\sum_{k\le n}m_k)$.
A Gaussian random walk is almost surely positive infinitely often: along
$n_j=2^{2^j}$, the endpoint of the previous block divided by $\sqrt{qn_j}$ tends
to zero almost surely (Chebyshev and Borel--Cantelli, since $n_{j-1}/n_j$ is
summable), while the normalized increments of the new blocks are independent
standard normals and exceed one infinitely often. Hence
$\sum_ke^{\theta_k}=\infty$ almost surely, and dominated convergence gives
$S_n\downarrow0$. A finite $\sum_km_k$ would bound $S_n$ away from zero.
\end{proof}

\begin{propS}[Exact CT mean]\label{prop:S-bayes}
Let $q_c>0$ and let $\theta_0$ be a fixed number or a Gaussian variable
independent of the Brownian motion. Put $z(x)=\sqrt{8e^x/q_c}$ and
$M_j=\E\int_0^\infty u^je^{-z(\theta_0)\cosh u}du$ for $j=0,2$. Then, as
$t\to\infty$,
\begin{equation}
\Pr(N_t=0)=\sqrt{\frac8{\pi q_ct}}\,M_0\Bigl(1+\frac{B_X}t+O(t^{-2})\Bigr),\qquad
\E[e^{\theta_t}\mid N_t=0]=\frac1{2t}+\frac{B_X}{t^2}+O(t^{-3}),\label{eq:S-bayes}
\end{equation}
with $B_X=\pi^2/(2q_c)-2M_2/(q_cM_0)$.
\end{propS}

\begin{proof}
By Brownian scaling,
$\int_0^te^{x+\sqrt{q_c}B_s}ds\overset d=(4e^x/q_c)\int_0^{q_ct/4}e^{2B_s}ds$.
The spectral (Alili--Gruet) representation~\cite[§4, p.~322]{Matsumoto2005}
gives
\[
S(t,x)=\frac2\pi\int_0^\infty e^{-q_ct\nu^2/8}\cosh(\pi\nu/2)K_{i\nu}(z(x))\,d\nu .
\]
With $K_{i\nu}(z)=\int_0^\infty e^{-z\cosh u}\cos(\nu u)du$ and
$|\cos v-1+v^2/2|\le v^4/24$,
$F(\nu)=\cosh(\pi\nu/2)\E K_{i\nu}(z)=M_0+A_2\nu^2+O(\nu^4)$ with
$A_2=(\pi^2M_0/4-M_2)/2$. All moments used are finite for Gaussian $\theta_0$,
since the $j$th integral grows only like $(1+|x|)^{j+1}$ as $x\to-\infty$, and
$|F(\nu)|\le M_0\cosh(\pi\nu/2)$. Splitting the $\nu$-integral at one and using
$\int_0^\infty\nu^{2i}e^{-s\nu^2}d\nu$ with $s=q_ct/8$ gives the survival
expansion. Differentiating the Cox likelihood gives
$-S'(t)=\E[e^{\theta_t}e^{-\int_0^te^{\theta_s}ds}]$, and differentiating under
the integral multiplies the integrand by $q_c\nu^2/8$. The same domination
applies, so $-S'(t)=\tfrac12\sqrt{8/(\pi q_c)}M_0t^{-3/2}(1+3B_X/t+O(t^{-2}))$.
The ratio $-S'/S$ is the conditional mean.
\end{proof}

Hence the ratio of the exact mean to the Gaussian median is asymptotically
$q_ct/12\to\infty$. The two quantities are different readouts, and
Fig.~1B of the main text compares readouts, not risks.
Table~\ref{tab:S-bayes} gives the numerical comparison, computed by the
spectral integral. For $q_c=0.002$/s at $t=1000$\,s, $t\,m(t)=2.85$, far from
$1/2$; the expansion is not uniform as $q_c\downarrow0$.

\begin{table}[ht]\centering\small
\caption{Exact CT Bayes mean, $\lambda_0=20$/s, $V_0=0.05$, $q_c=0.2$/s ($B_X=24.216$\,s).}
\label{tab:S-bayes}
\begin{tabular}{@{}rrrrr@{}}\toprule
$t$ (s) & exact mean & $1/(2t)$ & $1/(2t)+B_X/t^2$ & Gaussian median $6/(q_ct^2)$\\\midrule
20 & 0.080306 & 0.025 & 0.085541 & 0.075\\
60 & 0.014858 & 0.008333 & 0.015060 & 0.008333\\
100 & 0.0073775 & 0.005 & 0.0074216 & 0.003\\
500 & 0.0010965 & 0.001 & 0.0010969 & 0.00012\\
1000 & $5.2417\times10^{-4}$ & $5\times10^{-4}$ & $5.2422\times10^{-4}$ & $3\times10^{-5}$\\
10000 & $5.0242\times10^{-5}$ & $5\times10^{-5}$ & $5.0242\times10^{-5}$ & $3\times10^{-7}$\\
\bottomrule\end{tabular}
\end{table}

\section{Wake-up}\label{sec:wake}

This section proves Proposition~2 on the exact wake-up identity and its
distribution (Proposition~\ref{prop:S-wake}), proves Corollary~1, treats the
first nonzero bin and the crossing width, and bounds post-peak cooling under
bounded counts (Proposition~\ref{prop:S-cool}).

\subsection{Proof of Proposition 2}

\begin{propS}\label{prop:S-wake}
In the notation of Proposition~2, for any $C\ge0$,
\begin{equation}
\log(\Delta\lambda^+)=\log s+gC-d=\log s+vC-\frac{vs(vC+1)}{1+vs}.\label{eq:S-id}
\end{equation}
Consequently
\begin{equation}
0\le(\log s+vC)-\log(\Delta\lambda^+)\le vs(vC+1),\label{eq:S-onesided}
\end{equation}
and, for the first-order value $A_n(C)=\log\{6/(qn^2)\}+qnC/3$,
\begin{equation}
|\log(\Delta\lambda^+)-A_n(C)|\le\Bigl|\log\frac{sqn^2}6\Bigr|+C\Bigl|v-\frac{qn}3\Bigr|+\frac{vs(vC+1)}{1+vs}.
\label{eq:S-finite}
\end{equation}
If $C\sim\Pois(m)$ and the silent state is given, then $\lambda^+=Ae^{gC}$,
$\E(\lambda^+)^r=A^r\exp\{m(e^{rg}-1)\}$ for $r>0$,
\begin{equation}
\Pr(\lambda^+>R\Lambda)=\Pr\bigl(C>\log(R\Lambda/A)/g\bigr),\label{eq:S-tail}
\end{equation}
and the $p$-quantile of $\lambda^+$ is $Ae^{gQ_p(C)}$. If the next count is
independent $\Pois(m)$ and $\lambda^+$ is the plug-in prediction, the expected
excess loss of the next bin is
\begin{equation}
\E R_{\rm next}=se^{-d}\exp\{m(e^g-1)\}-m+m\{\log(m/s)+d-gm\}.\label{eq:S-Rnext}
\end{equation}
\end{propS}

\begin{proof}
The explicit update gives $V^+=g$ and $\theta^+=\theta_n+g(C-s)$; with
$g=v-v^2s/(1+vs)$ this is~\eqref{eq:S-id}. The bounds~\eqref{eq:S-onesided}
and~\eqref{eq:S-finite} follow from nonnegativity and the triangle inequality.
The map $C\mapsto Ae^{gC}$ is strictly increasing, so quantiles and tail
probabilities transfer from $C$; the moments follow from the Poisson generating
function. For~\eqref{eq:S-Rnext}, the conditional expected loss with prediction
$\mu=\Delta\lambda^+$ is $\mu-m+m\log(m/\mu)$; substitute
$\E\mu=se^{-d}\exp\{m(e^g-1)\}$ and $\E\log\mu=\log s-d+gm$.
\end{proof}

The bound~\eqref{eq:S-finite} has unit coefficients and needs only the
computable silent state. It is a check for a given state, not a bound in terms
of $(n,q)$ alone. ``Peak'' here means the value after the first wake-up
update; it bounds the window maximum from below, and later counts can create
new peaks.

\paragraph{Benchmark.} For $q=0.01$, $\Delta=0.05$\,s ($q_c=0.2$/s), $n=1200$
($L=60$\,s), initial rate 20/s, $V_0=0.05$, the silent state is
$v=4.074702$, $g=4.068034$, $\lambda^-=0.0080451$/s. With $C=50$,
$\log_{10}\lambda^+=86.2411$, while the first-order value is $84.780$. With
$C\sim\Pois(50)$, $\log_{10}$ of the median, the mean and the 95\% quantile of
$\lambda^+$ are $86.2411$, $1245.2408$ and $107.4417$. The mean is dominated
by counts near $me^g$, far in the Poisson tail; a few thousand Monte Carlo
paths do not estimate it.

\subsection{Proof of Corollary 1 and the crossing width}

At $C=m=\Lambda\Delta$, \eqref{eq:S-id} gives $\lambda^+>\Lambda$ if and only if
$\log s+gm-d>\log m$, that is $g\Lambda\Delta>\log(\Lambda/\lambda^-)+d$,
which is~(6). Replacing $m$ by the count quantile $c_{1-\alpha}$ gives the
level-$\alpha$ criterion $g\,c_{1-\alpha}\le\log(R\Lambda/\lambda^-)+d$ for
staying below $R\Lambda$ with probability at least $1-\alpha$, by~\eqref{eq:S-tail}.

\paragraph{First nonzero bin.} If every bin after the wake boundary has full
exposure and the ``wake bin'' is the first with $C\ge1$, the number $J$ of extra
zero bins and the first nonzero count satisfy
\begin{equation}
\Pr(J=j)=e^{-mj}(1-e^{-m}),\qquad
\Pr(C=c\mid C\ge1)=\frac{e^{-m}m^c/c!}{1-e^{-m}},\quad c\ge1,\label{eq:S-zt}
\end{equation}
and are independent. The update is applied to the state after $n+J$ zero
bins. Given $J=j$, the zero-truncated $p$-quantile is
$F^{-1}_{\Pois(m)}\{e^{-m}+p(1-e^{-m})\}$, to be inserted
in~\eqref{eq:S-id}. As $\Delta\to0$, the first nonzero count equals one with
probability tending to one, but the physical wait $J\Delta$ tends to an
$\mathrm{Exp}(\Lambda)$ time. Hence a small $v\E C$ does not imply a small
first-event update. If the wake occurs inside a bin with remaining exposure
$u\in(0,\Delta]$, all statements hold conditionally with $m$ replaced by
$\Lambda u$; the filter's own $s=\lambda^-\Delta$ is unchanged.

\paragraph{Unit exponent versus crossing.} With $q=q_c\Delta$, $n=L/\Delta$ and
the leading silence law, $v\approx q_cL/3$ and $\lambda^-\approx6/(q_cL^2)$. At
the mean count the exponent is $gC\approx q_cL\Lambda\Delta/3$, which equals one
at $\Delta_{\rm unit}=3/(q_cL\Lambda)$. Crossing requires the larger exponent
in~(6); neglecting $d$ and finite-silence corrections,
\begin{equation}
\Delta_{\rm cross}\approx\frac{3\log(q_cL^2\Lambda/6)}{q_cL\Lambda}.\label{eq:S-cross}
\end{equation}
A more accurate value freezes the CT silent state $(\lambda_L,W_L)$ and solves
the mean-input crossing for the bin width:
\begin{equation}
h=\log(\Lambda/\lambda_L),\qquad
\Delta_{\rm cross,frozen}=\frac{h}{W_L(\Lambda-\lambda_L-h\lambda_L)},\label{eq:S-frozen}
\end{equation}
valid when $h>0$ and the denominator is positive. These are approximations, not
a phase-transition theorem. For $q_c=0.2$/s, $L=60$\,s and $\Lambda=1000$/s,
$\Delta_{\rm unit}=0.25$\,ms (0.24541\,ms with the exact CT state) and
$\Delta_{\rm cross,frozen}=2.880$\,ms. The raw discrete recursion with mean input
is still below $\Lambda$ at 1\,ms and above it at 3\,ms. A scan over 81 cells
($\Delta$ from 0.1\,ms to 1\,s, $L\in\{5,20,60\}$\,s, $\Lambda\in\{10,100,1000\}$/s)
gave median absolute errors of the first-order $\log\lambda^+$ of 0.398, 0.108
and 0.037 for the three $L$, and maximum errors of 40--117 at coarse bins and
high counts.

\subsection{Cooling after a large peak}

\begin{propS}[Cooling]\label{prop:S-cool}
Number the bins from the post-wake peak and let $0\le C_j\le B$. For the
explicit filter,
\begin{equation}
\theta_{j+1}-\theta_j=-1+\varepsilon_j,\qquad
\varepsilon_j=\frac{1+(V_j+q)C_j}{1+(V_j+q)s_j}.\label{eq:S-cool}
\end{equation}
Let $M>B+q^{-1}$ and $s_0>M$. While $s_j>M$,
$0\le\varepsilon_j\le\eta:=(B+q^{-1})/M<1$. The number $N$ of steps until
$s_N\le M$ satisfies
$\lceil\log(s_0/M)\rceil\le N\le\lceil\log(s_0/M)/(1-\eta)\rceil$, and
\[
s_0\frac{1-e^{-N}}{1-e^{-1}}\le\sum_{j<N}s_j\le s_0\frac{1-e^{-(1-\eta)N}}{1-e^{-(1-\eta)}}.
\]
If $M\ge m$, the realized excess loss over these bins satisfies
$\sum_{j<N}s_j-N\{m+B\log(s_0/m)\}\le R_N\le\sum_{j<N}s_j$.
\end{propS}

\begin{proof}
\eqref{eq:S-cool} is a rearrangement of the explicit update, and
$V_j+q\ge q$ bounds $\varepsilon_j$. Each log decrement lies in $[1-\eta,1]$;
summing gives the bounds on $N$, and exponentiating gives the geometric sums.
In the high region $\log(s_j/m)\in[0,\log(s_0/m)]$, which bounds the loss terms.
\end{proof}

Poisson counts have no global bound, so the path bound is used with $B$ such that
$J\Pr\{\Pois(m)>B\}\le\alpha$ over a window of $J$ bins, which must contain the
whole descent. Under the deterministic mean input $C_j=m$ and a peak
$s_0\to\infty$, the number of bins to reach $(1+\delta)m$ is
$\log(s_0/m)+O(1)$, so $T_{\rm cool}=\Delta\log(\lambda_{\rm peak}/\Lambda)+O(\Delta)$.
If the scoring window $J\to\infty$ with $J\log s_0=o(s_0)$, then
\begin{equation}
R_J\sim\frac{s_0}{1-e^{-1}}=\lambda_{\rm peak}\Delta\frac e{e-1},\label{eq:S-coolloss}
\end{equation}
because $s_j/s_0\to e^{-j}$ for each fixed $j$, the high region is controlled by
the geometric tail, and the low region contributes at most $JM=o(s_0)$. For fixed
$J$ the constant is $(1-e^{-J})/(1-e^{-1})$. The $O(1)$ terms depend on
$q,m,\delta$ and are not uniform as $q\downarrow0$.

\section{The implicit update}\label{sec:implicit}

This section proves Lemma~1 and Proposition~3 and gives a bound uniform in the
bin width for a fixed silent state (Proposition~\ref{prop:S-fixedstate}). Here
$v=V^-$ is the prior variance, $\lambda=e^{\theta^-}$ the predicted rate, and
$s=\Delta\lambda$ the predicted count.

\subsection{Proof of Lemma 1}

The implicit equation is $f(u)=u-\log\lambda+v(\Delta e^u-C)=0$. Since
$f'(u)=1+v\Delta e^u>0$ and $f\to\pm\infty$ at $\pm\infty$, the root is unique.
For $C>0$, $f(\log\lambda)=v(s-C)$ and $f(\log(C/\Delta))=\log(C/s)$ have
opposite signs or both vanish, so $\lambda^+$ lies between $\lambda$ and
$C/\Delta$. For $C=0$, $\theta^+=\log\lambda-v\Delta e^{\theta^+}<\log\lambda$,
and $e^{\theta^+}\le\lambda$ gives $\theta^+\ge\log\lambda-vs$. This is the
scalar no-overshoot property of~\cite[Prop.~1]{Lange2026}; its Lambert-$W$ form
is~\cite[eq.~(C.1)]{Lange2026}. Over $H$ bins it gives
$\max_{j\le H}\lambda_j\le\max\{\lambda_0,\max_jC_j/\Delta\}$, a bound that does
not depend on the silence but does depend on the data and on $\Delta$.

\subsection{The wake-up expansion}

\begin{lemma}\label{lem:S-lw}
Let $C>s$, $k=vC$, $a=vs$, $h=\log(C/s)=\log(k/a)$, and suppose $0<h<k$. Then
\begin{equation}
1-\frac hk\le\frac{\lambda^+}{C/\Delta}\le1-\frac hk+\frac{-\log(1-h/k)}k,\qquad
0\le\theta^+-\Bigl[\log\frac C\Delta+\log\Bigl(1-\frac hk\Bigr)\Bigr]\le\frac{-\log(1-h/k)}{k-h}.
\label{eq:S-lw}
\end{equation}
\end{lemma}

\begin{proof}
With $w=v\Delta\lambda^+$, the implicit equation is $w+\log w=k+\log a=k+\log k-h$.
At $w_0=k-h$ the left minus the right side equals $\log(1-h/k)<0$, and at $w=k$
it equals $h>0$. The derivative is $1+1/w\ge1$, so
$0\le w-w_0\le-\log(1-h/k)$. Divide by $k$; the second bound follows from
$\log(1+x)\le x$.
\end{proof}

After a long silence, $h=\log(qCn^2/6)+o(1)$ and $k\sim qnC/3$, so $h/k\to0$ and
the implicit rate approaches $C/\Delta$. The gap $h$ is close to $\log(nk/2)$;
its leading term is $2\log n$. In the opposite regime $ae^k\ll1$,
$|\theta^+-(\theta^-+k-ae^k)|\le(ae^k)^2$, and the implicit rate can stay far below
the observed rate. In the benchmark of Appendix~\ref{sec:wake}, with $C=50$ and
the explicit silent state, the implicit update gives 942.71/s ($h=11.730$,
$k=203.735$); from the implicit filter's own silent state it gives 942.86/s.

\subsection{Proof of Proposition 3}

Before the first positive count $J$, finitely many further zero counts do not
change the leading terms of Proposition~\ref{prop:S-silence}: uniformly in
$j\le H$, the predicted state satisfies
$\theta^-_j=\log\{6/(q\Delta)\}-2\log n+o(1)$ and $v_j\sim qn/3$. The loss of a
bin with $C=0$ is $s_j-m$, which is bounded. If $J\le H$ and $C_J=c\ge1$, the
loss of bin $J$ is $s_J-m-c\log(s_J/m)=2c\log n+O(1)$. By
Lemma~\ref{lem:S-lw}, $h/k\to0$, so $\lambda_J\to c/\Delta$ and $V_J\to1/c$.
The remaining bins are finitely many continuous functions of state and counts.
On every fixed count path, the only divergent term of the window loss is
therefore $2C_J\log n$, and it is absent if the whole window is zero.

To exchange limit and expectation, note that uniformly in $c\ge1$,
\[
\frac{\log(c/s)}{vc}\le\frac{\log(1/s)+1/e}{v}\to0 .
\]
Hence for large $n$, after the first positive count, $\lambda_J\ge c/(2\Delta)$
and $V_J\le2/c\le2$. Later prior variances are at most $B_1=2+Hq$. By Lemma~1,
later rates are at most $N/\Delta$, where $N$ is the window count, and the log
rate is at least $-\log(2\Delta)-B_1HN$. After the divergent term is removed,
the absolute window loss is bounded by $K(1+N^2)$ independently of $n$. Since
$N\sim\Pois(Hm)$ has a finite second moment, dominated convergence applies.
Finally,
\[
\E[C_J\mathbf 1\{J\le H\}]=\sum_{j=1}^He^{-(j-1)m}\E C_j=\frac{m(1-e^{-Hm})}{1-e^{-m}},
\]
which gives~(7). The constant $O(1)$ depends on $q_c$, $\Delta$, $\Lambda$, $H$
and the initial state. The first wake bin is scored before its update, so a
posterior close to $C/\Delta$ cannot remove the $O(\log n)$ loss of the
preceding low prediction.

\subsection{A bound uniform in the bin width for a fixed silent state}

The next statement clarifies what the implicit update does guarantee as
$\Delta\downarrow0$: finiteness for a fixed silent state, with constants that
deteriorate with the silence.

\begin{propS}\label{prop:S-fixedstate}
Fix a post-silence state $(\lambda_0,V_0)$, a window $T=H\Delta$, its count
$N_T$, and $B=V_0+q_cT$. For any uniform binning of the same event path, the
implicit predicted rates satisfy $\max_j\lambda_j\le\bar A:=\lambda_0e^{BN_T}$. Under a
Poisson$(\Lambda)$ stream,
\[
\E\max_j\lambda_j\le\lambda_0\exp\{\Lambda T(e^B-1)\},\qquad
\E\mathcal R_T\le T(1+\Lambda BT)\lambda_0\exp\{\Lambda T(e^B-1)\}+\Lambda T\,|\log(\Lambda/\lambda_0)|.
\]
\end{propS}

\begin{proof}
Posterior variances do not exceed prior variances, so every prior variance is at
most $B$. The implicit mean equation gives
$\theta_j-\theta_{j-1}=v_j(C_j-\Delta\lambda_j^+)\le BC_j$; summing gives the
rate bound. It also gives $\theta_j-\theta_{j-1}\ge-B\Delta\bar A$, so every
predicted log rate is at least $\log\lambda_0-BT\bar A$. Summing~\eqref{eq:kl},
$\sum_jd_\Delta(\Lambda,\lambda_j)\le T(1+\Lambda BT)\bar A+\Lambda T|\log(\Lambda/\lambda_0)|$.
Take expectations with the Poisson moment generating function, and use that the
expected realized loss equals the expected sum of conditional divergences.
\end{proof}

If all binnings start from the same pre-silence state $(\lambda_i,V_i)$, summing
the $C=0$ case of Lemma~1 over the silence gives
$\lambda_i\exp\{-(V_i+q_cL)\lambda_iL\}\le\lambda_0\le\lambda_i$ and
$V_0\le V_i+q_cL$. The resulting bounds are uniform in $\Delta$ for fixed $L$ and
$T$, but they are not uniform in $L$.

\section{No joint bound: proof of Proposition 4}\label{sec:ct}

This section establishes fixed-path convergence of E and I to the
continuous-time filter (Lemma~\ref{lem:S-conv}), bounds the decay after the
initial event (Lemma~\ref{lem:S-decay}), and constructs the expected-risk
counterexample that proves Proposition~4.

\subsection{Fixed-path convergence to the CT filter}

\begin{lemma}\label{lem:S-conv}
Fix $q_c>0$, a finite horizon $L+T$, a finite initial state, and an event path
with finitely many distinct event times, none at time zero. Bin it at width
$\Delta$ with $q=q_c\Delta$. As $\Delta\downarrow0$, both the E and the I paths
converge to the CT filter: uniformly on closed intervals avoiding the event
times, and in the Skorokhod $J_1$ topology on the whole window after each event
is moved to the right end of its bin. On this path the error is $O(\Delta)$.
\end{lemma}

\begin{proof}
For small $\Delta$, each bin contains at most one event. By
$W\le W_0+q_c(L+T)$ and the CT path bound
$\lambda_{\max}\le\lambda_0\exp\{N(W_0+q_c(L+T))\}$, the CT path stays in a
compact set. On it, a zero-count E step satisfies
$\Delta\theta=-We^\theta\Delta+O(\Delta^2)$ and
$\Delta W=(q_c-W^2e^\theta)\Delta+O(\Delta^2)$ uniformly, and the implicit
function theorem gives the same expansion for I. In a bin with one event, E gives
$\theta^+=\theta^-+W^-+O(\Delta)$, $W^+=W^-+O(\Delta)$. For I,
$\theta^+=\theta^-+(W+q_c\Delta)(1-\Delta e^{\theta^+})$ together with
$\theta^+\le\theta^-+W+q_c\Delta$ gives the same limit. The flow and jump maps are
locally Lipschitz near the compact set. A discrete Gronwall argument on each
event-free interval, combined with the finitely many jumps, gives an $O(\Delta)$
error by induction; for small $\Delta$ this also keeps the discrete path in the
chosen neighborhood. Moving events to bin ends shifts times by at most $\Delta$.
\end{proof}

The CT limit itself is the filter of~\cite[§6]{Eden2008}. Because E and I share
it, the implicit update cannot avoid overshoot for every silence length.

\subsection{Decay after the first event}

After a silence of length $L$ from a fixed initial state, let $(\lambda_0,W_0)$
be the CT state. A single event gives $\lambda_A=\lambda_0e^{W_0}$ and
$W_A=W_0$. By Proposition~\ref{prop:S-ct}, $W_0=q_cL/3+\beta+o(1)$ and
$\lambda_0\sim6/(q_cL^2)$ for a constant $\beta$, so
\begin{equation}
\lambda_A\sim\frac{6e^\beta}{q_cL^2}\exp(q_cL/3).\label{eq:S-lamA}
\end{equation}
Even for a fixed $L$ and $W_0>0$, there is no deterministic peak bound
depending only on $W_0$ and $\Lambda$: $M$ events packed into a short interval
raise the log rate by nearly $MW_0$, and such clusters have positive
probability.

\begin{lemma}\label{lem:S-decay}
After the event, while no further event occurs, $W^2=a/\lambda+K_A\lambda^2$ with
$a=2q_c/3$ and $K_A=(W_A^2\lambda_A-a)/\lambda_A^3$. Suppose $K_A>0$, and let
$\lambda_c=(a/K_A)^{1/3}$ and $R=\lambda_A/\lambda_c\ge2$, which is equivalent to
$W_A^2\lambda_A\ge9a$. The time $\tau_c$ to decay from $\lambda_A$ to
$\lambda_c$ and the integrated rate $J_c=\int_0^{\tau_c}\lambda\,dt$ satisfy
\begin{equation}
\tau_c\le\frac{K_A^{1/6}}{2a^{2/3}},\qquad J_c\ge\frac1{2\sqrt2\,a^{1/3}K_A^{1/6}}.\label{eq:S-tauJ}
\end{equation}
If moreover $\lambda_c\ge6\Lambda$, then
$\int_0^{\tau_c}d(\Lambda,\lambda(t))\,dt\ge1/(4\sqrt2\,a^{1/3}K_A^{1/6})$, where
$d(\Lambda,\lambda)=\lambda-\Lambda+\Lambda\log(\Lambda/\lambda)$.
\end{lemma}

\begin{proof}
The invariant is Proposition~\ref{prop:S-ct} started at the event. On
$\lambda\ge\lambda_c$, $a/\lambda\le K_A\lambda_c^2\le K_A\lambda^2$, so
$\sqrt{K_A}\lambda\le W\le\sqrt{2K_A}\lambda$. Substituting in
$dt=-d\lambda/(W\lambda^2)$ gives
$\tau_c\le(\lambda_c^{-2}-\lambda_A^{-2})/(2\sqrt{K_A})\le K_A^{1/6}/(2a^{2/3})$,
and substituting in $\lambda\,dt=-d\lambda/(W\lambda)$ gives
$J_c\ge(\lambda_c^{-1}-\lambda_A^{-1})/\sqrt{2K_A}\ge1/(2\sqrt2\lambda_c\sqrt{K_A})$
when $R\ge2$. For $x=\lambda/\Lambda\ge6$, $x/2\ge1+\log x$, so
$d(\Lambda,\lambda)\ge\lambda/2$ on the whole descent.
\end{proof}

For long silences $K_A\sim W_A^2/\lambda_A^2\downarrow0$, so the descent time
tends to zero while, by~\eqref{eq:S-lamA}, the divergence integral is at least
\begin{equation}
\frac{3e^{\beta/3}}{4\sqrt2\,q_cL}\exp(q_cL/9)\{1+o(1)\}.\label{eq:S-lower}
\end{equation}

\subsection{From a path to an expected-risk counterexample}

Let a Poisson$(\Lambda)$ stream resume at the end of the silence, fix $h>0$, and
score a window of length $2h$. Let $A_h$ be the event of exactly one event in
the first $h$ and none in the second; $\Pr(A_h)=\Lambda he^{-2\Lambda h}$. On
$A_h$ the filter is silent until $L+S$ with $S\in[0,h]$. The asymptotics above
are uniform over this compact range, so for large $L$, $R\ge2$,
$\lambda_c\ge6\Lambda$ and $\tau_c\le h$ hold for every $S$. Since the CT filter
obeys an envelope of the type of Proposition~\ref{prop:S-fixedstate} for each
fixed $L$, the expected realized loss equals the expected divergence integral
over the full probability space. The integrand is nonnegative, so restricting to
$A_h$ gives
\begin{equation}
\E\mathcal R^{\rm CT}_{2h}\ge\Lambda he^{-2\Lambda h}\min_{0\le S\le h}
\frac1{4\sqrt2\,a^{1/3}K_A(L+S)^{1/6}}=\Omega\Bigl(\frac{e^{q_cL/9}}{q_cL}\Bigr).
\label{eq:S-ctrisk}
\end{equation}
With $h=1/\Lambda$ the event probability is $e^{-2}$.

For each fixed $L$, Lemma~\ref{lem:S-conv} gives pathwise convergence of the
implicit filter, and of its piecewise constant predicted rates in integral, to
the CT filter. Fatou's lemma applied to the nonnegative divergence gives
\[
\E\mathcal R^{\rm CT}_{2h}\le\liminf_{\Delta\downarrow0}\E\mathcal R^{I,\Delta}_{2h}.
\]
If all silences and windows must be whole numbers of bins, take $h=1/\Lambda$,
$L_k=kh$ and $\Delta_i=h/i$, let $i\to\infty$ for each $k$, then let
$k\to\infty$. Hence any bound on $\E\mathcal R^{I,\Delta}_{2h}$ valid for all
small $\Delta$ must grow at least like $e^{q_cL/9}/(q_cL)$ in $L$. In particular
there is no $O(\log L)$ bound uniform in $\Delta$, and no uniform constant. This
does not contradict Proposition~3, which fixes $\Delta$: the two limits do not
commute. The same argument applies to E.

\begin{table}[ht]\centering\small
\caption{CT first event immediately after a silence of length $L$ ($q_c=0.2$/s,
initial rate 20/s, $W_i=0.05$). The divergence integral is exact for the descent to
$\lambda_c$ at $\Lambda=1000$/s. The first two rows do not satisfy
$\lambda_c\ge6\Lambda$, so~\eqref{eq:S-tauJ} is not used there.}
\label{tab:S-ct}
\begin{tabular}{@{}rrrr@{}}\toprule
$L$ (s) & post-event rate (/s) & time to $\lambda_c$ (s) & divergence integral (nats)\\\midrule
240 & $4.9414\times10^3$ & 0.2440 & 552.65\\
360 & $6.5671\times10^6$ & 0.025392 & 56.78\\
600 & $2.1060\times10^{13}$ & $2.0407\times10^{-4}$ & $1.4366\times10^4$\\
1000 & $2.8947\times10^{24}$ & $4.6871\times10^{-8}$ & $6.2557\times10^7$\\
\bottomrule\end{tabular}
\end{table}

\section{Experimental protocol and deviations}\label{sec:protocol}

This section documents the prespecified protocol, its amendments and engineering
corrections, the numerical implementations, the synthetic and real-data designs,
and the statistical tests.

\subsection{Protocol freezing and amendments}

A SHA-256 protocol hash bound the protocol, code, binaries, frozen parameters,
and theory regions, with final value
\texttt{2e74c40575a6f5d0cd0be585b08dc61f}\allowbreak\texttt{48ea4921109e139fc89af5268949e29e}.
Formal runs required a separate human approval file carrying the same hash and a
full pass of ten gates: build; per-bin C++ against a NumPy reference (absolute
tolerance $10^{-12}$, relative $10^{-13}$; log10 of extended losses within
$10^{-12}$); theory regressions; causality; binning; approval guard; statistical
fixtures; source and seed checks; resources; and a formal-path smoke test.

\paragraph{Disclosure.} Descriptive summaries of the real test days had been
inspected before the experimental design was finalized. These summaries included
the distribution of inter-trade gaps, trade counts in the first second after
wake-ups, and a price spike on 2026-10-17 at 21:22 UTC. The design thus reflects these test-day
summaries. No filter had been run on the test days, and no parameter,
threshold, sample size or analysis was chosen from the formal results.

\paragraph{Amendment 1 (before formal test evaluation).} The parameter grid was
widened once for all methods after E selected the lower edge $q_c=10^{-4}$ of
the original grid on the training days at 250\,ms and 1000\,ms, the bin widths
of R2 and R1. The grid went from $q_c\in10^{\{-4,-3.5,\dots,1\}}$ to
$10^{\{-6,-5.5,\dots,2\}}$, $V_{\max}$ from $\{0.25,0.5,1,2\}$ to
$\{0.0625,\dots,2\}$, and $\kappa_{\rm OU}$ to include 1/s (154 to 323
candidates), and all training tasks were rerun in full. Thereafter, E selected
$q_c=10^{-4}$ as an interior point with its training gap to I unchanged;
remaining edge selections are marked in Table~\ref{tab:frozen}.

\paragraph{Preflight corrections (before formal test evaluation).}
Preflight updates were recorded under new hashes while preserving earlier
hashes, code, and records: (i) the X scoring window was aligned with the
Gaussian window (SW from 60\,s, CTRL from 300\,s, replacing a 60\,s CTRL start);
and (ii) Holm-adjusted confirmation fields were added for R1--R3 with unchanged
thresholds.

\paragraph{Engineering repair after the first formal attempt.}
The first approved formal run filled the disk by retaining kernel input and
output files. The engine was then updated to delete them after validation, and
a disk gate was added. The scientific design and frozen parameters were
unchanged; the repair was recorded under a new hash with the earlier records
preserved.

\paragraph{Execution.} Attempt 1 (earlier hash) stopped with a full disk after
all 1312 synthetic checkpoints and no real-data unit. Attempt 2 (final hash)
stopped after 26\,s with a Windows sharing violation when an external monitor
read a process log during file replacement, leaving 17 checkpoints. Attempt 3
resumed on the final hash without the monitor and completed in 3213\,s (27\% of
the frozen extrapolation of 11865\,s), with exit code 0, empty standard error,
and 1312/1312 synthetic and 20/20 real units. At load time, the SHA-256 of each
of the 20 real files matched its published CHECKSUM.

\subsection{Methods and numerics}

\begin{itemize}
\item E and I are~(2) and~(3). E is evaluated with a log-domain gain. I is solved in
closed form with the Wright omega function: with
$z=\log(v\Delta)+\theta^-+vC$, $w+\log w=z$ and
$\theta^+=\log w-\log(v\Delta)$, switching to $\theta^-+vC-w$ for $z<-2$,
without unguarded Newton iteration or clipping of $\theta$ or scored rates.
\item E$_{\rm cap}$ and I$_{\rm cap}$ replace $v$ by $\min(V+q,V_{\max})$ before the
update; $V_{\max}=1$ in synthetic data, tuned on real data.
\item OU$_{\rm I}$ (real data only) predicts with
$\theta^-=\mu+e^{-\kappa\Delta}(\theta-\mu)$, $v=e^{-2\kappa\Delta}V+q$, then applies
the implicit update.
\item X is an uncertified numerical grid-Bayes approximation for the same
random-walk prior on $[-200,16]$ with step $\min(0.025,\sqrt q/8)$, direct
convolution with a kernel truncated at $\pm32$ standard deviations, and scoring
by the full Poisson mixture. Across all 2624 X stream records, the largest mass
within 5 log units of the lower edge was $3.5\times10^{-34}$ and the largest
per-step boundary loss was $7.2\times10^{-15}$.
\item Scores and losses are stored as extended numbers $(\mathrm{sign},\log_{10}|x|)$,
so legal overflow of E is scored exactly instead of being discarded.
\end{itemize}

\subsection{Synthetic streams}

The synthetic benchmark comprises SW and CTRL stream families generated with
seeds disjoint from all gate and tuning seeds. SW streams have 300\,s at rate
$\Lambda$, a true silence of length $L$, and 120\,s at $\Lambda$; scoring drops
the first 60\,s and includes the silence. Counts per constant-rate segment are
Poisson with conditionally uniform times, and the same continuous event stream
is binned at all five widths and used for both $q_c$. Filters start at
$\theta=\log\Lambda$ with $V$ set to the stationary E variance, the positive
root of $V^2+qV=q/(\Lambda\Delta)$. CTRL streams last 1800\,s and are scored
after 300\,s. Their base log rate is $\tfrac12\log(1000)[1-\cos(\pi t/150)]$,
plus a three-state jump process on $\{-0.5,0,+0.5\}$ with exponential holding
times of mean 12.5\,s, generated by thinning; their oracle integrates the
intensity over each bin by five-point Gauss--Legendre quadrature, split at
jumps.

\paragraph{Regions.} Each cell's $\kappa$ is a cell-level design prediction
computed from the deterministic E recursion over $n=L/\Delta$ zero counts from
the initial state with mean count $m$, omitting the random 300\,s pre-phase.
Table~\ref{tab:kappa} lists all 200 values. At fixed $\Delta$, $\kappa$ is
exactly invariant under $(q_c,\Lambda)\mapsto(10q_c,\Lambda/10)$ because the
initial variance scales by ten, the zero-count recursion depends only on $V/q$
and $qs$, and $g\Lambda\Delta$, $\Lambda/\lambda^-$, and $d$ remain unchanged;
this invariance produces the repeated rows of the table while the realized
streams of such cells still differ.

\subsection{Tests}

Synthetic families H1, H2 and H3 each use level $0.05/3$ with Holm correction
across cells and strict $p_{\rm Holm}<0.05/3$.
\begin{itemize}
\item H1: in each crossing cell, loss(I) $<$ loss(E); exact one-sided sign test on 64
paired streams, ties removed and reported.
\item H2: in each cell with $\Delta\in\{1,10\}$\,ms and $q_cL/3\ge8$ (24 cells),
both I$_{\rm cap}<$ I and E$_{\rm cap}<$ E; an intersection--union test whose $p$ is
the larger of the two sign-test $p$ values.
\item H3: in each CTRL cell ($\Delta,q_c$), both $\mathrm{loss(I)}-1.02\,\mathrm{loss(E)}<0$
and $\mathrm{loss(I_{cap})}-1.02\,\mathrm{loss(I)}<0$, by paired $t$ tests combined
as an intersection--union test.
\end{itemize}
Real data use level 0.05 with Holm correction across R1--R3. R1 and R2 are exact
one-sided sign tests of score(I) $>$ score(E) over 20 contract-days at 1000\,ms
and 250\,ms. R3 tests at 1000\,ms whether the mean of
$d=\mathrm{score(I)}-\mathrm{score(I_{cap})}-0.005|\mathrm{score(I)}|$ is negative
(paired $t$, 19 df). The absolute value fixes the sign convention for a
percentage of a negative log score. ``Not confirmed'' means that the test did not
reach its Holm-adjusted level, so the null hypothesis was retained; it is not
evidence of equivalence. Power was stated analytically before the run: the
strictest first Holm level is $0.05/(3\times73)$ for H1 and $0.05/(3\times24)$
for H2, and CTRL (32 streams) and real data (20 pairs) have limited power.

\subsection{Real data}

The real-data evaluation uses Binance USD-M USDT perpetual \texttt{aggTrades}
for SNDK, MU, TSLA, NVDA, SPY, SPX, QQQ, MSTR, COIN, and
AAPL~\cite{BinanceVision} (CC BY-NC-SA 4.0; raw data are not redistributed).
Each aggregate trade is one event at its millisecond transaction time, with
records sharing a millisecond kept. Each day is filtered independently from
00:00 UTC with a one-hour burn-in. Training days were 2026-10-10/11 and test
days 2026-10-17/18. For each method and $\Delta\in\{10,50,250,1000\}$\,ms, one
parameter set shared by all contracts maximized the total training log score;
every candidate was run in full on every training day, with ties broken by the
smallest $q_c$, $V_{\max}$, $\kappa$. All filters start at $\theta_0=\mu$, the
equal-weight average over contracts of the log mean training rate, and $V_0=1$. Table~\ref{tab:frozen} lists the selected
parameters.

\subsection{Descriptive definitions fixed before reporting}

The preflight review found five descriptive quantities ambiguous; they were
defined as follows before reporting.
(i) Quantiles of the first wake-up update are reported descriptively by theory
region without a dedicated prespecified test.
(ii) The ``window maximum'' is taken over the full scoring window (SW $[60,T)$,
CTRL $[300,1800)$\,s).
(iii) Recovery is the delay from the wake-up to the start of the first
one-second run in which the prediction stays within $[1/1.5,1.5]$ times the true
bin-average rate; streams that never recover are counted separately.
(iv) The ``maximum one-second count'' is tallied over fixed UTC seconds rather
than a sliding window.
(v) Because CTRL streams contain no wake-up transition, their wake fields hold
sentinel values and are omitted from wake-up summaries.

\section{Additional results}\label{sec:results}

This section reports confirmatory outcomes and descriptive summaries, all read
from the frozen results file and checkpoints; no analysis was rerun or
reselected. Pooled records share event streams
across bin widths and $q_c$ rather than forming independent samples. Quantiles
are nearest-rank; means are on the original scale and are dominated by extreme
streams.

\subsection{Synthetic confirmatory detail}

H1 is confirmed in all 73 crossing cells (Table~\ref{tab:h1}); 67 of them are
unanimous, and the largest Holm-adjusted $p$ is $2.0\times10^{-8}$. H2 is confirmed
in 2 of 24 cells (Table~\ref{tab:h2}). Over all 24 cells, I$_{\rm cap}$ beats I in 200
and loses in 1336 of 1536 paired comparisons, while E$_{\rm cap}$ beats E in 1203 and
loses in 333. Because the stored summaries record whole-window losses, first-update
values, window maxima, and recovery times without per-bin traces or a phase
split, the extra loss of I$_{\rm cap}$ is measured over the full window rather than
separated between silence and wake-up. H3 is confirmed in all 10 CTRL cells
(Table~\ref{tab:h3}).

\begin{table}[ht]\centering\small
\caption{H3 per CTRL cell (32 streams): mean (SE) of
$d_1=\mathrm{loss(I)}-1.02\,\mathrm{loss(E)}$ and
$d_2=\mathrm{loss(I_{cap})}-1.02\,\mathrm{loss(I)}$, in nats, and the Holm-adjusted
intersection--union $p$.}
\label{tab:h3}
\begin{tabular}{@{}rrrrr@{}}\toprule
$\Delta$ (ms) & $q_c$ & $d_1$ & $d_2$ & $p_{\rm Holm}$\\\midrule
1 & 0.02 & $-23.59$ (0.29) & $-23.59$ (0.29) & $4.0\times10^{-37}$\\
1 & 0.2 & $-40.18$ (0.21) & $-39.98$ (0.21) & $2.1\times10^{-48}$\\
10 & 0.02 & $-23.75$ (0.30) & $-23.77$ (0.30) & $5.4\times10^{-37}$\\
10 & 0.2 & $-42.01$ (0.23) & $-40.05$ (0.22) & $4.7\times10^{-48}$\\
50 & 0.02 & $-24.66$ (0.35) & $-24.78$ (0.34) & $3.4\times10^{-35}$\\
50 & 0.2 & $-48.67$ (0.54) & $-40.05$ (0.26) & $1.4\times10^{-38}$\\
250 & 0.02 & $-30.12$ (1.11) & $-30.07$ (0.53) & $7.9\times10^{-23}$\\
250 & 0.2 & $-67.23$ (3.18) & $-36.34$ (0.42) & $8.8\times10^{-20}$\\
1000 & 0.02 & $-87.29$ (8.53) & $-57.97$ (1.60) & $9.2\times10^{-12}$\\
1000 & 0.2 & $-133.78$ (10.33) & $-56.82$ (1.57) & $4.8\times10^{-14}$\\
\bottomrule\end{tabular}
\end{table}

\subsection{Regions and wake-up behavior}

\begin{table}[ht]\centering\small
\caption{Excess loss by theory region, pooled over cells and streams (C: 4672,
N: 5888, T: 2240 records per method).}
\label{tab:regions}
\begin{tabular}{@{}llrrrrr@{}}\toprule
Region & Method & Q1 & Median & Q3 & Mean & $>1000$ nats\\\midrule
C & E & $1.3\times10^{11}$ & $5.9\times10^{40}$ & $3.1\times10^{237}$ & $10^{37018}$ & 97.2\%\\
C & I & 488 & 1356 & 2467 & 2662 & 56.4\%\\
C & E$_{\rm cap}$ & 761 & $4.5\times10^{7}$ & $5.2\times10^{39}$ & $10^{459}$ & 74.4\%\\
C & I$_{\rm cap}$ & 514 & 1244 & 2464 & 2499 & 56.2\%\\\addlinespace
N & E & 98 & 307 & 1089 & $6.1\times10^{60}$ & 28.1\%\\
N & I & 57 & 145 & 369 & 320 & 12.2\%\\
N & E$_{\rm cap}$ & 61 & 148 & 375 & 329 & 13.7\%\\
N & I$_{\rm cap}$ & 61 & 148 & 375 & 329 & 13.7\%\\\addlinespace
T & E & 27 & 67 & 166 & $5.4\times10^{18}$ & 9.0\%\\
T & I & 27 & 55 & 149 & 149 & 2.9\%\\
T & E$_{\rm cap}$ & 27 & 63 & 156 & 157 & 3.0\%\\
T & I$_{\rm cap}$ & 27 & 60 & 150 & 151 & 2.9\%\\
\bottomrule\end{tabular}
\end{table}

The no-crossing region classifies cells, not individual streams: in it, E has
a median excess loss of 307\,nats but a mean of
$6.1\times10^{60}$\,nats. Part of this mean comes from the cells with $m\ll1$
discussed in the main text, where the zero-truncated count rather than the mean
count predicts the crossing.

\begin{table}[ht]\centering\small
\caption{Wake-up behavior by region: share of streams whose first post-wake
prediction of E exceeds $\Lambda$ and $10\Lambda$; median first update and median
window maximum of the predicted rate (/s); recovery (recovered/total and median
delay among recovered streams).}
\label{tab:wake}
\begin{tabular}{@{}llrrrrr@{}}\toprule
Region & Method & first $>\Lambda$ & first $>10\Lambda$ & first update & window max & recovery\\\midrule
C & E & 92.96\% & 91.67\% & $8.9\times10^{39}$ & $2.2\times10^{41}$ & 3282/4672, 9\,s\\
C & I & & & 123 & 202 & 4672/4672, 0.25\,s\\
C & E$_{\rm cap}$ & & & $1.3\times10^{8}$ & $1.5\times10^{8}$ & 4224/4672, 3.5\,s\\
C & I$_{\rm cap}$ & & & 86 & 186 & 4672/4672, 0.5\,s\\\addlinespace
N & E & 1.89\% & 1.22\% & 0.053 & 134 & 5888/5888, 1.75\,s\\
N & I & & & 0.053 & 36 & 5888/5888, 1.21\,s\\\addlinespace
T & E & 13.13\% & 4.60\% & 0.73 & 14.5 & 2240/2240, 3\,s\\
T & I & & & 0.69 & 5.7 & 2240/2240, 3\,s\\
\bottomrule\end{tabular}
\end{table}

\subsection{Logarithmic implicit risk}

Table~\ref{tab:slope} compares the growth of the median implicit loss with $L$
at $\Lambda=1000$/s and $\Delta=1$\,s against the increment $2m\log(L_2/L_1)$ of
Proposition~3, with $m=1000$. The window factor
$(1-e^{-Hm})/(1-e^{-m})$ equals one to machine precision. Whereas Proposition~3
concerns expectations as $n\to\infty$, the table compares cell medians at finite
$n$.

\begin{table}[ht]\centering\small
\caption{Increase of the median implicit loss (nats) between successive silence
lengths, $\Lambda=1000$/s, $\Delta=1$\,s.}
\label{tab:slope}
\begin{tabular}{@{}lrrr@{}}\toprule
$L_1\to L_2$ (s) & $q_c=0.02$ & $q_c=0.2$ & $2m\log(L_2/L_1)$\\\midrule
$5\to20$ & 2617 & 2719 & 2773\\
$20\to60$ & 2117 & 2129 & 2197\\
$60\to300$ & 3142 & 3142 & 3219\\
$300\to1200$ & 2728 & 2722 & 2773\\
$60\to1200$ & 5870 & 5864 & 5991\\
\bottomrule\end{tabular}
\end{table}

\subsection{Exact Bayes at 1~s}

Table~\ref{tab:xgap} lists the median paired gap between X and I. The gap grows
with both $\Lambda$ and $L$, consistent with the Gaussian median readout falling
far below the exact predictive mean after a silence (Appendix~\ref{sec:silence}).
Because X is computed only at this synthetic bin width, the comparison is
restricted to this setting rather than finer bins or real data.

\subsection{Real data}

Table~\ref{tab:real} gives the totals. At 1000\,ms, I exceeds E by 64,228 nats
in total log score, 2.48\% of the absolute total of E, and OU$_{\rm I}$ has the
highest total. E produces 13 large-rate bins (1 at 10\,ms, 12 at 50\,ms) and I
none; E$_{\rm cap}$ produces 17 (12 at 50\,ms, 5 at 250\,ms), and at 250\,ms its
total is the lowest of all methods. I$_{\rm cap}$ has a higher total than I at
10\,ms and a lower one at 50, 250, and 1000\,ms; R3 is unconfirmed ($p=0.91$;
mean of $d$ 2277\,nats, SE 1657). Because real data lack an oracle, the post-gap
column reports the negative log score directly rather than an excess loss.

\paragraph{Dependence across contract-days (exploratory).} The confirmatory
tests R1--R3 treat the 20 contract-days as exchangeable pairs, as the protocol
specified. The ten contracts trade on the same two calendar days, so market-wide
conditions can make their score differences dependent, and the two days of one
contract can be serially dependent. As a check outside the protocol, computed
from the frozen per-contract-day scores without rerunning any filter, each
contract's I$-$E score difference was summed over its two test days and an exact
one-sided sign test was applied to the ten sums. I beat E for 10 of 10 contracts
at 1000\,ms ($p=9.8\times10^{-4}$) and 9 of 10 at 250\,ms ($p=0.011$); the
exception at 250\,ms is SNDK, which favored E on both test days. Per day, I beat E
on 8 and 10 of the ten contracts at 1000\,ms and on 9 and 9 at 250\,ms. At the
descriptive widths, I beat E on 20 of 20 contract-days at 10\,ms and 18 of 20 at
50\,ms. With only two test days, day-level clustering cannot be tested.

\clearpage
\input{supp_tables}

\clearpage
\section*{Acknowledgment}
Claude Code with the Claude Opus 5.5 model was used to assist with implementing
the experimental code described in Section~\ref{sec:exp}.
The authors take full responsibility for the experimental code and the content
of this article.

\section*{Data and Code Availability}
The experimental protocol, code, frozen parameters, and results are available from the
authors. The exchange data are from Binance Vision \cite{BinanceVision}, licensed
under CC BY-NC-SA 4.0.

\end{document}

%% file: supp_tables.tex
\begin{longtable}{@{}rrrrrrrr@{}}
\caption{Frozen crossing index $\kappa$ of all 200 SW cells (C: crossing, $\kappa\ge\log10$; N: no crossing, $\kappa\le-\log10$; T: transition).}\label{tab:kappa}\\
\toprule $q_c$ (1/s) & $\Lambda$ (1/s) & $L$ (s) & \multicolumn{5}{c}{$\kappa$ at $\Delta$ = 1, 10, 50, 250, 1000 ms}\\ \midrule\endfirsthead
\toprule $q_c$ & $\Lambda$ & $L$ & 1 & 10 & 50 & 250 & 1000\\ \midrule\endhead
\bottomrule\endfoot
0.02 & 1 & 5 & $-$0.555\,T & $-$0.554\,T & $-$0.551\,T & $-$0.537\,T & $-$0.484\,T\\
0.02 & 1 & 20 & $-$1.57\,T & $-$1.57\,T & $-$1.56\,T & $-$1.52\,T & $-$1.36\,T\\
0.02 & 1 & 60 & $-$3.01\,N & $-$3.01\,N & $-$2.99\,N & $-$2.88\,N & $-$2.51\,N\\
0.02 & 1 & 300 & $-$5.82\,N & $-$5.8\,N & $-$5.71\,N & $-$5.29\,N & $-$3.71\,N\\
0.02 & 1 & 1200 & $-$8.5\,N & $-$8.43\,N & $-$8.1\,N & $-$6.48\,N & $-$0.402\,T\\
\addlinespace
0.02 & 10 & 5 & $-$1.34\,T & $-$1.33\,T & $-$1.31\,T & $-$1.2\,T & $-$0.824\,T\\
0.02 & 10 & 20 & $-$3.09\,N & $-$3.08\,N & $-$3.01\,N & $-$2.68\,N & $-$1.51\,T\\
0.02 & 10 & 60 & $-$4.97\,N & $-$4.93\,N & $-$4.75\,N & $-$3.88\,N & $-$0.716\,T\\
0.02 & 10 & 300 & $-$8.02\,N & $-$7.84\,N & $-$7.03\,N & $-$2.96\,N & 12.2\,C\\
0.02 & 10 & 1200 & $-$10.7\,N & $-$9.98\,N & $-$6.77\,N & 9.29\,C & 69.4\,C\\
\addlinespace
0.02 & 100 & 5 & $-$2.73\,N & $-$2.69\,N & $-$2.52\,N & $-$1.69\,T & 0.682\,T\\
0.02 & 100 & 20 & $-$5.05\,N & $-$4.92\,N & $-$4.34\,N & $-$1.51\,T & 8.01\,C\\
0.02 & 100 & 60 & $-$7.11\,N & $-$6.74\,N & $-$5.09\,N & 3.06\,C & 32.3\,C\\
0.02 & 100 & 300 & $-$10.1\,N & $-$8.31\,N & $-$0.265\,T & 39.9\,C & 189\,C\\
0.02 & 100 & 1200 & $-$12.3\,N & $-$5.07\,N & 27\,C & 187\,C & 786\,C\\
\addlinespace
0.02 & 1000 & 5 & $-$4.6\,N & $-$4.27\,N & $-$2.81\,N & 3.81\,C & 14.9\,C\\
0.02 & 1000 & 20 & $-$7.11\,N & $-$5.88\,N & $-$0.432\,T & 26\,C & 106\,C\\
0.02 & 1000 & 60 & $-$9.01\,N & $-$5.38\,N & 10.7\,C & 90.3\,C & 367\,C\\
0.02 & 1000 & 300 & $-$10.6\,N & 7.42\,C & 87.5\,C & 487\,C & 1960\,C\\
0.02 & 1000 & 1200 & $-$7.38\,N & 64.7\,C & 385\,C & 1984\,C & 7953\,C\\
\addlinespace
0.2 & 1 & 5 & $-$1.34\,T & $-$1.33\,T & $-$1.31\,T & $-$1.2\,T & $-$0.824\,T\\
0.2 & 1 & 20 & $-$3.09\,N & $-$3.08\,N & $-$3.01\,N & $-$2.68\,N & $-$1.51\,T\\
0.2 & 1 & 60 & $-$4.97\,N & $-$4.93\,N & $-$4.75\,N & $-$3.88\,N & $-$0.716\,T\\
0.2 & 1 & 300 & $-$8.02\,N & $-$7.84\,N & $-$7.03\,N & $-$2.96\,N & 12.2\,C\\
0.2 & 1 & 1200 & $-$10.7\,N & $-$9.98\,N & $-$6.77\,N & 9.29\,C & 69.4\,C\\
\addlinespace
0.2 & 10 & 5 & $-$2.73\,N & $-$2.69\,N & $-$2.52\,N & $-$1.69\,T & 0.682\,T\\
0.2 & 10 & 20 & $-$5.05\,N & $-$4.92\,N & $-$4.34\,N & $-$1.51\,T & 8.01\,C\\
0.2 & 10 & 60 & $-$7.11\,N & $-$6.74\,N & $-$5.09\,N & 3.06\,C & 32.3\,C\\
0.2 & 10 & 300 & $-$10.1\,N & $-$8.31\,N & $-$0.265\,T & 39.9\,C & 189\,C\\
0.2 & 10 & 1200 & $-$12.3\,N & $-$5.07\,N & 27\,C & 187\,C & 786\,C\\
\addlinespace
0.2 & 100 & 5 & $-$4.6\,N & $-$4.27\,N & $-$2.81\,N & 3.81\,C & 14.9\,C\\
0.2 & 100 & 20 & $-$7.11\,N & $-$5.88\,N & $-$0.432\,T & 26\,C & 106\,C\\
0.2 & 100 & 60 & $-$9.01\,N & $-$5.38\,N & 10.7\,C & 90.3\,C & 367\,C\\
0.2 & 100 & 300 & $-$10.6\,N & 7.42\,C & 87.5\,C & 487\,C & 1960\,C\\
0.2 & 100 & 1200 & $-$7.38\,N & 64.7\,C & 385\,C & 1984\,C & 7953\,C\\
\addlinespace
0.2 & 1000 & 5 & $-$6.45\,N & $-$3.36\,N & 10.1\,C & 66.2\,C & 69.1\,C\\
0.2 & 1000 & 20 & $-$8.17\,N & 3.92\,C & 57.3\,C & 311\,C & 974\,C\\
0.2 & 1000 & 60 & $-$7.69\,N & 28.4\,C & 188\,C & 973\,C & 3601\,C\\
0.2 & 1000 & 300 & 5.1\,C & 185\,C & 985\,C & 4967\,C & 19552\,C\\
0.2 & 1000 & 1200 & 62.3\,C & 782\,C & 3982\,C & 19962\,C & 79512\,C\\
\addlinespace
\end{longtable}

\begin{longtable}{@{}lrrrrcrr@{}}
\caption{H1 per crossing cell: wins/losses/ties of I against E over 64 paired streams, Holm-adjusted $p$, and cell median excess loss (nats).}\label{tab:h1}\\
\toprule Cell & $L$ & $\Lambda$ & $q_c$ & $\Delta$ (ms) & W/L/T & $p_{\rm Holm}$ & median E / I\\ \midrule\endfirsthead
\toprule Cell & $L$ & $\Lambda$ & $q_c$ & $\Delta$ & W/L/T & $p_{\rm Holm}$ & median E / I\\ \midrule\endhead
\bottomrule\endfoot
SW2 & 5 & 100 & 0.2 & 250 & 64/0/0 & $4.0{\times}10^{-18}$ & 1896 / 508\\
SW2 & 5 & 100 & 0.2 & 1000 & 64/0/0 & $4.0{\times}10^{-18}$ & $2.4{\times}10^{8}$ / 652\\
SW3 & 5 & 1000 & 0.02 & 250 & 64/0/0 & $4.0{\times}10^{-18}$ & $1.5{\times}10^{4}$ / 1913\\
SW3 & 5 & 1000 & 0.02 & 1000 & 64/0/0 & $4.0{\times}10^{-18}$ & $4.7{\times}10^{9}$ / 5133\\
SW3 & 5 & 1000 & 0.2 & 50 & 64/0/0 & $4.0{\times}10^{-18}$ & $1.3{\times}10^{6}$ / 1678\\
SW3 & 5 & 1000 & 0.2 & 250 & 64/0/0 & $4.0{\times}10^{-18}$ & $1.1{\times}10^{31}$ / 2441\\
SW3 & 5 & 1000 & 0.2 & 1000 & 64/0/0 & $4.0{\times}10^{-18}$ & $1.7{\times}10^{33}$ / 7138\\
SW5 & 20 & 10 & 0.2 & 1000 & 64/0/0 & $4.0{\times}10^{-18}$ & $1.7{\times}10^{5}$ / 171\\
SW6 & 20 & 100 & 0.02 & 1000 & 64/0/0 & $4.0{\times}10^{-18}$ & $5.1{\times}10^{5}$ / 726\\
SW6 & 20 & 100 & 0.2 & 250 & 64/0/0 & $4.0{\times}10^{-18}$ & $7.0{\times}10^{12}$ / 574\\
SW6 & 20 & 100 & 0.2 & 1000 & 64/0/0 & $4.0{\times}10^{-18}$ & $2.5{\times}10^{48}$ / 918\\
SW7 & 20 & 1000 & 0.02 & 250 & 64/0/0 & $4.0{\times}10^{-18}$ & $5.0{\times}10^{13}$ / 2551\\
SW7 & 20 & 1000 & 0.02 & 1000 & 64/0/0 & $4.0{\times}10^{-18}$ & $5.2{\times}10^{49}$ / 7749\\
SW7 & 20 & 1000 & 0.2 & 10 & 54/10/0 & $2.0{\times}10^{-8}$ & 1732 / 1525\\
SW7 & 20 & 1000 & 0.2 & 50 & 64/0/0 & $4.0{\times}10^{-18}$ & $1.6{\times}10^{26}$ / 1797\\
SW7 & 20 & 1000 & 0.2 & 250 & 64/0/0 & $4.0{\times}10^{-18}$ & $10^{136}$ / 3069\\
SW7 & 20 & 1000 & 0.2 & 1000 & 64/0/0 & $4.0{\times}10^{-18}$ & $10^{429}$ / 9858\\
SW9 & 60 & 10 & 0.2 & 250 & 63/1/0 & $2.1{\times}10^{-17}$ & 795 / 165\\
SW9 & 60 & 10 & 0.2 & 1000 & 64/0/0 & $4.0{\times}10^{-18}$ & $1.5{\times}10^{15}$ / 188\\
SW10 & 60 & 100 & 0.02 & 250 & 54/10/0 & $2.0{\times}10^{-8}$ & 1526 / 466\\
SW10 & 60 & 100 & 0.02 & 1000 & 64/0/0 & $4.0{\times}10^{-18}$ & $3.2{\times}10^{16}$ / 953\\
SW10 & 60 & 100 & 0.2 & 50 & 63/1/0 & $2.1{\times}10^{-17}$ & $3.6{\times}10^{5}$ / 498\\
SW10 & 60 & 100 & 0.2 & 250 & 64/0/0 & $4.0{\times}10^{-18}$ & $3.0{\times}10^{42}$ / 630\\
SW10 & 60 & 100 & 0.2 & 1000 & 64/0/0 & $4.0{\times}10^{-18}$ & $10^{165}$ / 1142\\
SW11 & 60 & 1000 & 0.02 & 50 & 64/0/0 & $4.0{\times}10^{-18}$ & $3.6{\times}10^{6}$ / 1385\\
SW11 & 60 & 1000 & 0.02 & 250 & 64/0/0 & $4.0{\times}10^{-18}$ & $6.0{\times}10^{41}$ / 3086\\
SW11 & 60 & 1000 & 0.02 & 1000 & 64/0/0 & $4.0{\times}10^{-18}$ & $10^{163}$ / 9866\\
SW11 & 60 & 1000 & 0.2 & 10 & 64/0/0 & $4.0{\times}10^{-18}$ & $3.4{\times}10^{13}$ / 1556\\
SW11 & 60 & 1000 & 0.2 & 50 & 64/0/0 & $4.0{\times}10^{-18}$ & $4.8{\times}10^{83}$ / 1956\\
SW11 & 60 & 1000 & 0.2 & 250 & 64/0/0 & $4.0{\times}10^{-18}$ & $10^{425}$ / 3624\\
SW11 & 60 & 1000 & 0.2 & 1000 & 64/0/0 & $4.0{\times}10^{-18}$ & $10^{1572}$ / $1.2{\times}10^{4}$\\
SW12 & 300 & 1 & 0.2 & 1000 & 64/0/0 & $4.0{\times}10^{-18}$ & $4.9{\times}10^{5}$ / 57\\
SW13 & 300 & 10 & 0.02 & 1000 & 64/0/0 & $4.0{\times}10^{-18}$ & $3.0{\times}10^{6}$ / 165\\
SW13 & 300 & 10 & 0.2 & 250 & 64/0/0 & $4.0{\times}10^{-18}$ & $3.6{\times}10^{13}$ / 165\\
SW13 & 300 & 10 & 0.2 & 1000 & 64/0/0 & $4.0{\times}10^{-18}$ & $1.2{\times}10^{83}$ / 213\\
SW14 & 300 & 100 & 0.02 & 250 & 64/0/0 & $4.0{\times}10^{-18}$ & $8.0{\times}10^{18}$ / 535\\
SW14 & 300 & 100 & 0.02 & 1000 & 64/0/0 & $4.0{\times}10^{-18}$ & $1.4{\times}10^{84}$ / 1257\\
SW14 & 300 & 100 & 0.2 & 10 & 64/0/0 & $4.0{\times}10^{-18}$ & 3861 / 475\\
SW14 & 300 & 100 & 0.2 & 50 & 64/0/0 & $4.0{\times}10^{-18}$ & $1.6{\times}10^{30}$ / 507\\
SW14 & 300 & 100 & 0.2 & 250 & 64/0/0 & $4.0{\times}10^{-18}$ & $10^{213}$ / 712\\
SW14 & 300 & 100 & 0.2 & 1000 & 64/0/0 & $4.0{\times}10^{-18}$ & $10^{853}$ / 1448\\
SW15 & 300 & 1000 & 0.02 & 10 & 60/4/0 & $1.1{\times}10^{-13}$ & $3.5{\times}10^{4}$ / 1060\\
SW15 & 300 & 1000 & 0.02 & 50 & 64/0/0 & $4.0{\times}10^{-18}$ & $6.0{\times}10^{40}$ / 1532\\
SW15 & 300 & 1000 & 0.02 & 250 & 64/0/0 & $4.0{\times}10^{-18}$ & $10^{214}$ / 3902\\
SW15 & 300 & 1000 & 0.02 & 1000 & 64/0/0 & $4.0{\times}10^{-18}$ & $10^{853}$ / $1.3{\times}10^{4}$\\
SW15 & 300 & 1000 & 0.2 & 1 & 64/0/0 & $4.0{\times}10^{-18}$ & 2271 / 1467\\
SW15 & 300 & 1000 & 0.2 & 10 & 64/0/0 & $4.0{\times}10^{-18}$ & $4.2{\times}10^{81}$ / 1575\\
SW15 & 300 & 1000 & 0.2 & 50 & 64/0/0 & $4.0{\times}10^{-18}$ & $10^{438}$ / 2095\\
SW15 & 300 & 1000 & 0.2 & 250 & 64/0/0 & $4.0{\times}10^{-18}$ & $10^{2160}$ / 4441\\
SW15 & 300 & 1000 & 0.2 & 1000 & 64/0/0 & $4.0{\times}10^{-18}$ & $10^{8487}$ / $1.5{\times}10^{4}$\\
SW16 & 1200 & 1 & 0.2 & 250 & 64/0/0 & $4.0{\times}10^{-18}$ & $6.7{\times}10^{29}$ / 57\\
SW16 & 1200 & 1 & 0.2 & 1000 & 64/0/0 & $4.0{\times}10^{-18}$ & $2.5{\times}10^{30}$ / 59\\
SW17 & 1200 & 10 & 0.02 & 250 & 61/3/0 & $9.5{\times}10^{-15}$ & $2.4{\times}10^{6}$ / 121\\
SW17 & 1200 & 10 & 0.02 & 1000 & 64/0/0 & $4.0{\times}10^{-18}$ & $2.2{\times}10^{31}$ / 195\\
SW17 & 1200 & 10 & 0.2 & 50 & 64/0/0 & $4.0{\times}10^{-18}$ & $1.0{\times}10^{29}$ / 159\\
SW17 & 1200 & 10 & 0.2 & 250 & 64/0/0 & $4.0{\times}10^{-18}$ & $1.7{\times}10^{99}$ / 180\\
SW17 & 1200 & 10 & 0.2 & 1000 & 64/0/0 & $4.0{\times}10^{-18}$ & $10^{342}$ / 252\\
SW18 & 1200 & 100 & 0.02 & 50 & 64/0/0 & $4.0{\times}10^{-18}$ & $1.4{\times}10^{9}$ / 355\\
SW18 & 1200 & 100 & 0.02 & 250 & 64/0/0 & $4.0{\times}10^{-18}$ & $2.3{\times}10^{79}$ / 606\\
SW18 & 1200 & 100 & 0.02 & 1000 & 64/0/0 & $4.0{\times}10^{-18}$ & $10^{343}$ / 1522\\
SW18 & 1200 & 100 & 0.2 & 10 & 64/0/0 & $4.0{\times}10^{-18}$ & $1.9{\times}10^{28}$ / 479\\
SW18 & 1200 & 100 & 0.2 & 50 & 64/0/0 & $4.0{\times}10^{-18}$ & $10^{133}$ / 524\\
SW18 & 1200 & 100 & 0.2 & 250 & 64/0/0 & $4.0{\times}10^{-18}$ & $10^{828}$ / 772\\
SW18 & 1200 & 100 & 0.2 & 1000 & 64/0/0 & $4.0{\times}10^{-18}$ & $10^{3456}$ / 1713\\
SW19 & 1200 & 1000 & 0.02 & 10 & 64/0/0 & $4.0{\times}10^{-18}$ & $1.9{\times}10^{29}$ / 1075\\
SW19 & 1200 & 1000 & 0.02 & 50 & 64/0/0 & $4.0{\times}10^{-18}$ & $10^{176}$ / 1697\\
SW19 & 1200 & 1000 & 0.02 & 250 & 64/0/0 & $4.0{\times}10^{-18}$ & $10^{847}$ / 4545\\
SW19 & 1200 & 1000 & 0.02 & 1000 & 64/0/0 & $4.0{\times}10^{-18}$ & $10^{3447}$ / $1.6{\times}10^{4}$\\
SW19 & 1200 & 1000 & 0.2 & 1 & 64/0/0 & $4.0{\times}10^{-18}$ & $1.8{\times}10^{27}$ / 1467\\
SW19 & 1200 & 1000 & 0.2 & 10 & 64/0/0 & $4.0{\times}10^{-18}$ & $10^{341}$ / 1602\\
SW19 & 1200 & 1000 & 0.2 & 50 & 64/0/0 & $4.0{\times}10^{-18}$ & $10^{1801}$ / 2260\\
SW19 & 1200 & 1000 & 0.2 & 250 & 64/0/0 & $4.0{\times}10^{-18}$ & $10^{8499}$ / 5071\\
SW19 & 1200 & 1000 & 0.2 & 1000 & 64/0/0 & $4.0{\times}10^{-18}$ & $10^{34431}$ / $1.8{\times}10^{4}$\\
\end{longtable}

\begin{table}[ht]\centering\small
\caption{H2 per cell: wins/losses/ties of each component over 64 paired streams; IUT $p$ is the larger component $p$, Holm-adjusted across the 24 cells.}\label{tab:h2}
\begin{tabular}{@{}lrrrrccr@{}}\toprule
Cell & $L$ & $\Lambda$ & $q_c$ & $\Delta$ (ms) & I$_{\rm cap}<$I & E$_{\rm cap}<$E & $p_{\rm Holm}$\\ \midrule
SW12 & 300 & 1 & 0.2 & 1 & 55/9/0 & 64/0/0 & $4.1{\times}10^{-8}$\\
SW12 & 300 & 1 & 0.2 & 10 & 2/62/0 & 64/0/0 & 1\\
SW13 & 300 & 10 & 0.2 & 1 & 0/64/0 & 64/0/0 & 1\\
SW13 & 300 & 10 & 0.2 & 10 & 0/64/0 & 64/0/0 & 1\\
SW14 & 300 & 100 & 0.2 & 1 & 0/64/0 & 64/0/0 & 1\\
SW14 & 300 & 100 & 0.2 & 10 & 0/64/0 & 64/0/0 & 1\\
SW15 & 300 & 1000 & 0.2 & 1 & 0/64/0 & 64/0/0 & 1\\
SW15 & 300 & 1000 & 0.2 & 10 & 28/36/0 & 64/0/0 & 1\\
SW16 & 1200 & 1 & 0.02 & 1 & 1/63/0 & 4/60/0 & 1\\
SW16 & 1200 & 1 & 0.02 & 10 & 0/64/0 & 4/60/0 & 1\\
SW16 & 1200 & 1 & 0.2 & 1 & 64/0/0 & 64/0/0 & $1.3{\times}10^{-18}$\\
SW16 & 1200 & 1 & 0.2 & 10 & 5/59/0 & 64/0/0 & 1\\
SW17 & 1200 & 10 & 0.02 & 1 & 0/64/0 & 11/53/0 & 1\\
SW17 & 1200 & 10 & 0.02 & 10 & 0/64/0 & 26/38/0 & 1\\
SW17 & 1200 & 10 & 0.2 & 1 & 0/64/0 & 64/0/0 & 1\\
SW17 & 1200 & 10 & 0.2 & 10 & 0/64/0 & 64/0/0 & 1\\
SW18 & 1200 & 100 & 0.02 & 1 & 0/64/0 & 2/62/0 & 1\\
SW18 & 1200 & 100 & 0.02 & 10 & 0/64/0 & 40/24/0 & 1\\
SW18 & 1200 & 100 & 0.2 & 1 & 0/64/0 & 64/0/0 & 1\\
SW18 & 1200 & 100 & 0.2 & 10 & 0/64/0 & 64/0/0 & 1\\
SW19 & 1200 & 1000 & 0.02 & 1 & 0/64/0 & 28/36/0 & 1\\
SW19 & 1200 & 1000 & 0.02 & 10 & 16/48/0 & 64/0/0 & 1\\
SW19 & 1200 & 1000 & 0.2 & 1 & 0/64/0 & 64/0/0 & 1\\
SW19 & 1200 & 1000 & 0.2 & 10 & 29/35/0 & 64/0/0 & 1\\
\bottomrule\end{tabular}\end{table}

\begin{table}[ht]\centering\small
\caption{Median paired difference loss(X) $-$ loss(I) per cell at $\Delta=1$\,s (nats; negative favours exact Bayes). CTRL: -2044 ($q_c=0.02$) and -1450 ($q_c=0.2$).}\label{tab:xgap}
\begin{tabular}{@{}rrrrrrr@{}}\toprule
$q_c$ & $\Lambda$ & $L=5$ & 20 & 60 & 300 & 1200\\ \midrule
0.02 & 1 & $-$0.4978 & $-$1.239 & $-$1.103 & 1.577 & 4.489\\
0.02 & 10 & $-$7.24 & $-$14.07 & $-$16.65 & $-$25.68 & $-$52.67\\
0.02 & 100 & $-$165.3 & $-$338.6 & $-$536.1 & $-$832.3 & $-$1089\\
0.02 & 1000 & $-$4142 & $-$6661 & $-$8733 & $-$1.186e+04 & $-$1.459e+04\\
\addlinespace
0.2 & 1 & $-$0.7225 & $-$1.971 & $-$2.706 & $-$4.369 & $-$4.901\\
0.2 & 10 & 7.106 & $-$11.83 & $-$24.39 & $-$45.71 & $-$83.69\\
0.2 & 100 & $-$194.8 & $-$446.9 & $-$673.5 & $-$983 & $-$1238\\
0.2 & 1000 & $-$6218 & $-$8936 & $-$1.105e+04 & $-$1.418e+04 & $-$1.692e+04\\
\addlinespace
\bottomrule\end{tabular}\end{table}

\begin{table}[ht]\centering\small
\caption{Parameters frozen on the training days: $q_c$ (1/s); V$_{\max}$ for capped filters; $\kappa_{\rm OU}$ (1/s) for OU$_{\rm I}$. $^\ast$Selection on the edge of the revised grid. All filters start at $\theta_0=\mu=-1.6726$, $V_0=1$.}\label{tab:frozen}
\begin{tabular}{@{}rlllll@{}}\toprule
$\Delta$ (ms) & E & I & E$_{\rm cap}$ & I$_{\rm cap}$ & OU$_{\rm I}$\\ \midrule
10 & $10^{-3.0}$ & $10^{-2.5}$ & $10^{-3.0}$, 0.0625$^\ast$ & $10^{2.0}$, 0.5$^\ast$ & $10^{-1.0}$, $10^{-1}$\\
50 & $10^{-3.5}$ & $10^{-2.5}$ & $10^{-3.5}$, 0.0625$^\ast$ & $10^{1.0}$, 0.25 & $10^{-2.0}$, $10^{-2}$\\
250 & $10^{-4.0}$ & $10^{-3.0}$ & $10^{-3.5}$, 0.0625$^\ast$ & $10^{0.0}$, 0.125 & $10^{-2.5}$, $10^{-2}$\\
1000 & $10^{-4.0}$ & $10^{-3.0}$ & $10^{-4.0}$, 0.125 & $10^{-1.5}$, 0.0625$^\ast$ & $10^{-3.0}$, $10^{-3}$\\
\bottomrule\end{tabular}\end{table}

\begin{table}[ht]\centering\small
\caption{Real data, 20 test contract-days: total log score (higher is better), negative log score in the 5\,s after gaps of at least 20\,s, and bins whose predicted rate exceeds 100 times the day maximum of the fixed-second count.}\label{tab:real}
\begin{tabular}{@{}rlrrr@{}}\toprule
$\Delta$ (ms) & Method & Total log score & Post-gap 5\,s & Large-rate bins\\ \midrule
10 & E & $-$8951778 & 264261 & 1\\
10 & I & $-$8920922 & 267549 & 0\\
10 & E$_{\rm cap}$ & $-$8948392 & 264297 & 0\\
10 & I$_{\rm cap}$ & $-$8876917 & 278077 & 0\\
10 & OU$_{\rm I}$ & $-$8979223 & 260519 & 0\\
\addlinespace
50 & E & $-$6602464 & 205720 & 12\\
50 & I & $-$6443261 & 212440 & 0\\
50 & E$_{\rm cap}$ & $-$6596778 & 205789 & 12\\
50 & I$_{\rm cap}$ & $-$6586208 & 219287 & 0\\
50 & OU$_{\rm I}$ & $-$6454560 & 212253 & 0\\
\addlinespace
250 & E & $-$4135224 & 147671 & 0\\
250 & I & $-$4108589 & 152454 & 0\\
250 & E$_{\rm cap}$ & $-$4712659 & 149627 & 5\\
250 & I$_{\rm cap}$ & $-$4218511 & 158171 & 0\\
250 & OU$_{\rm I}$ & $-$4139487 & 151049 & 0\\
\addlinespace
1000 & E & $-$2591436 & 103896 & 0\\
1000 & I & $-$2527208 & 109084 & 0\\
1000 & E$_{\rm cap}$ & $-$2591436 & 103896 & 0\\
1000 & I$_{\rm cap}$ & $-$2585391 & 109778 & 0\\
1000 & OU$_{\rm I}$ & $-$2524346 & 108074 & 0\\
\bottomrule\end{tabular}\end{table}